\documentclass{article}

\usepackage[pdftex,
pdfauthor={Anirban Chaudhuri, Alexandre Marques, Karen Willcox},
pdftitle={mfEGRA: Multifidelity Efficient Global Reliability Analysis through Active Learning for Failure Boundary Location},
pdfkeywords={multi-fidelity, active learning, adaptive sampling, probability of failure, contour location, classification, Gaussian process, kriging, multiple information sources, EGRA, surrogate},
pdfproducer={Latex with hyperref},
pdfcreator={pdflatex}]{hyperref}

\newenvironment{keywords}%
{\begin{trivlist}\item[]{\bfseries Keywords:}\ }% oder "Keywords:"
	{\end{trivlist}}

\usepackage[margin=1in]{geometry} % Wider paper-format
\usepackage{graphicx}
\usepackage{wrapfig}% embedding figures/tables in text (i.e., Galileo style)
\usepackage{threeparttable}% tables with footnotes
\usepackage{multirow}
\usepackage{dcolumn}% decimal-aligned tabular math columns
\newcolumntype{d}{D{.}{.}{-1}}
\usepackage{subfigure}% subcaptions for subfigures
\usepackage{subfigmat}% matrices of similar subfigures, aka small multiples
\usepackage{lettrine}% dropped capital at beginning of paragraph
\usepackage{hyperref}% embedding hyperlinks [must be loaded after dropping]

\usepackage{amsfonts}
\usepackage{amsmath}
\usepackage{amssymb}

\usepackage{algorithm}
\usepackage{algpseudocode}

\usepackage{booktabs}
\usepackage{color}
\usepackage{url}

\usepackage{bbm}

\usepackage[inline]{enumitem}

\usepackage{tikz}
\usetikzlibrary{shapes,arrows,shapes.multipart,calc,backgrounds,fit,arrows.meta}
\usetikzlibrary{positioning}

\title{mfEGRA: Multifidelity Efficient Global Reliability Analysis through Active Learning for Failure Boundary Location}

\author{%
	Anirban Chaudhuri\footnote{Postdoctoral Associate, Department of Aeronautics and Astronautics, anirbanc@mit.edu.}, \ 
	Alexandre N. Marques\footnote{Postdoctoral Associate, Department of Aeronautics and Astronautics, noll@mit.edu.} \\
	{\normalsize\itshape Massachusetts Institute of Technology, Cambridge, MA, 02139, USA} \\
	Karen E. Willcox\footnote{Director, Oden Institute for Computational Engineering and Sciences, kwillcox@oden.utexas.edu}  \\ {\normalsize\itshape University of Texas at Austin, Austin, TX, 78712, USA}
}
\date{}

\newcommand{\failset}{\mathcal{G}}

\newcommand{\bz}{\boldsymbol{z}}

\newcommand{\hg}{\widehat{g}}
\newcommand{\GP}{\operatorname{GP}}

\newcommand{\IG}{\text{IG}}
\newcommand{\tP}{\text{P}}
\newcommand{\tF}{\text{F}}
\newcommand{\pr}{\text{pr}} % prior
\newcommand{\argmax}{\operatornamewithlimits{arg\ max}}

\newcolumntype{L}[1]{>{\raggedright\let\newline\\\arraybackslash\hspace{0pt}}m{#1}}
\newcolumntype{C}[1]{>{\centering\let\newline\\\arraybackslash\hspace{0pt}}m{#1}}

\renewcommand{\algorithmicrequire}{\textbf{Input:}}
\renewcommand{\algorithmicensure}{\textbf{Output:}}
\algrenewcommand\textproc{}

\graphicspath{{figures_MFEGRApaper/}}

\begin{document}
\maketitle

% % % % % % % % % % % % % % % % % % % % % % % %
\begin{abstract}

This paper develops mfEGRA, a multifidelity active learning method using data-driven adaptively refined surrogates for failure boundary location in reliability analysis. This work addresses the issue of prohibitive cost of reliability analysis using Monte Carlo sampling for expensive-to-evaluate high-fidelity models by using cheaper-to-evaluate approximations of the high-fidelity model. The method builds on the Efficient Global Reliability Analysis (EGRA) method, which is a surrogate-based method that uses adaptive sampling for refining Gaussian process surrogates for failure boundary location using a single-fidelity model. 
%This paper proposes a new method that combines information from multiple models with different fidelity to improve computational efficiency for adaptively refining the surrogate for accurately predicting the failure boundary. 
Our method introduces a two-stage adaptive sampling criterion that uses a multifidelity Gaussian process surrogate to leverage multiple information sources with different fidelities. The method combines expected feasibility criterion from EGRA with one-step lookahead information gain to refine the surrogate around the failure boundary. The computational savings from mfEGRA depends on the discrepancy between the different models, and the relative cost of evaluating the different models as compared to the high-fidelity model. We show that accurate estimation of reliability using mfEGRA leads to computational savings of $\sim$46\% for an analytic multimodal test problem and 24\% for a three-dimensional acoustic horn problem, when compared to single-fidelity EGRA. We also show the effect of using \textit{a priori} drawn Monte Carlo samples in the implementation for the acoustic horn problem, where mfEGRA leads to computational savings of 45\% for the three-dimensional case and 48\% for a rarer event four-dimensional case as compared to single-fidelity EGRA.
\end{abstract}


% % % % % % % % % % % % % % % % % % % % % % % %
\section{Introduction}
The presence of uncertainties in the manufacturing and operation of systems make reliability analysis critical for system safety. The reliability analysis of a system requires estimating the probability of failure, which can be computationally prohibitive when the high fidelity model is expensive to evaluate. In this work, we develop a method for efficient reliability estimation by leveraging multiple sources of information with different fidelities to build a multifidelity approximation for the limit state function. 

\begin{figure}[!h]
	\centering
	\begin{tikzpicture}
	\tikzstyle{block} = [rectangle, draw, fill=blue!15!white, text width=7em, text centered, rounded corners, minimum height=0.05em, inner ysep=0.1cm, inner xsep=0.05cm,font=\fontsize{8}{9.6}\selectfont]
	\tikzstyle{line} = [draw, thick, color=black!60, -latex']
	%\tikzstyle{line1} = [draw={rgb,255:red,235; green,130; blue,10},line width=15pt,-{Latex[scale=0.5]}]
	\tikzstyle{line1} = [draw=gray!30,line width=15pt,-{Latex[scale=0.5]}]
	%\tikzstyle{line1} = [draw=red!60!black,line width=5pt,-latex']
	%\tikzset{line1/.style={single arrow, draw, minimum width=1mm, minimum height=3mm,inner sep=0mm, single arrow head extend=1mm}}
	\tikzstyle{container} = [draw, very thick, black!70, rectangle, inner xsep=0.1cm, inner ysep=0.1cm, rounded corners,fill=gray!15]
	
	%%%%%%%%%%%%%%%%%%%%%%%%%%%%%%%%%%%%%%%%%%%%%%%%%%%
	% Reliability analysis using HF
	\node (a) at (0,0)
	{
		\begin{tikzpicture}[scale=1, auto]	
		% Place nodes
		\node [block] (Inner) {Reliability analysis loop};
		\node [block, below=2cm of Inner] (HF) {High-fidelity\\ model};	
		% Draw edges
		\path [line] (Inner.east) -| ++ (0.25,0) |- node[near start, font=\fontsize{8}{7}\selectfont,black] {\rotatebox{-90}{Random variable realization}} (HF.east);
		\path [line] (HF.west) -| ++ (-0.25,0) |- node[near start, font=\fontsize{8}{7}\selectfont,black] {\rotatebox{90}{System outputs}} (Inner.west);	
		\end{tikzpicture}
	};
	%%%%%%%%%%%%%%%%%%%%%%%%%%%%%%%%%%%%%%%%%%%%%%%%%%%
	% Reliability analysis using adaptive surrogates
	\node (b) [right=-0.2cm of a.east]
	{
		\begin{tikzpicture}[scale=1, auto]
		% Place nodes
		\node [block, text width=8.5em] (Inner) {Reliability analysis loop};
		\node [block, below=2cm of Inner, text width=8.5em] (ALS) {Single fidelity adaptive surrogate};
		% Draw edges
		\path [line] (Inner.east) -| ++ (0.25,0) |- node[near start, font=\fontsize{8}{7}\selectfont,black] {\rotatebox{-90}{Random variable realization}} (ALS.east);
		\path [line] (ALS.west) -| ++ (-0.25,0) |- node[near start, font=\fontsize{8}{7}\selectfont,black] {\rotatebox{90}{System outputs}} (Inner.west);
		\end{tikzpicture}
	};
	%%%%%%%%%%%%%%%%%%%%%%%%%%%%%%%%%%%%%%%%%%%%%%%%%%%
	% Multifidelity reliability analysis
	\node (c) [right=-0.2cm of b.east]
	{
		\begin{tikzpicture}[scale=1, auto]	
		% Place nodes
		\node [block, text width=8.5em] (Inner) {Reliability analysis loop};
		\node [block, below=1.1cm of Inner, text width=9.2em] (HF) {High-fidelity model};
		\node [block, below=0.1cm of HF, text width=9.2em] (LF1) {Low-fidelity model 1};
		\node [block, below=0.5cm of LF1, text width=9.2em] (LFk) {Low-fidelity model k};
		\path (LF1.south) -- node[auto=false,yshift=0.1cm,xshift=-0.1cm,font=\fontsize{6}{7}\selectfont]{\textbf{\vdots}} (LFk.north);
		% Draw box enclosing outputstats method
		\begin{scope}[on background layer]
		\node [container, label={[font=\fontsize{8}{7.5}\selectfont, black!70,label distance=-0.1cm,text width=3.5cm,align=center]90:{\textbf{Multifidelity\\ adaptive surrogate}}}, fit=(HF) (LF1) (LFk)] (M1) {};
		\end{scope}		
		% Draw edges
		\path [line] (Inner.east) -| ++ (0.45,0) |- node[near start, font=\fontsize{8}{7}\selectfont,black] {\rotatebox{-90}{Random variable realization}} ($(M1.north east)!0.7!(M1.south east)$);
		\path [line] ($(M1.north west)!0.7!(M1.south west)$) -| ++ (-0.2,0) |- node[near start, font=\fontsize{8}{7}\selectfont,black] {\rotatebox{90}{System outputs}} (Inner.west);	
		\end{tikzpicture}
	};
	%\draw (a.south east) node {(a)};
	\draw (a.south)+(0cm,-0.25cm) node {(a)};
	\draw (b.south)+(0cm,-0.25cm) node {(b)};
	\draw (c.south)+(0cm,-0.15cm) node {(c)};
	%\path [line1] (a.south)+(0,-0.5cm) -- (a.south-|c.south);
	\path [line1] (a.north)+(-1cm,0.5cm) -- node[black!90]{\normalsize\textbf{Improving computational efficiency}} ++(9.5cm,0.5cm);
\end{tikzpicture}
\caption{Reliability analysis with (a) high-fidelity model, (b) single fidelity adaptive surrogate, and (c) multifidelity adaptive surrogate.}
\label{fig:reliabilityAna}
\end{figure}
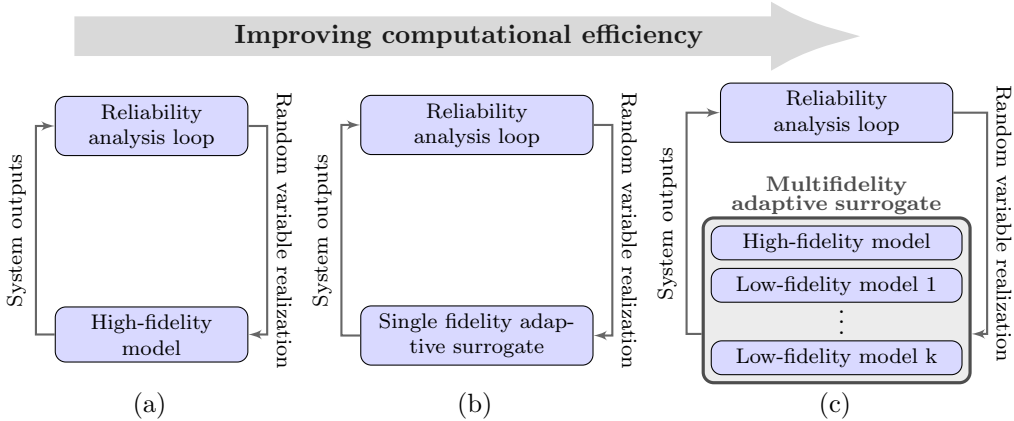

Reliability analysis for strongly non-linear systems typically require Monte Carlo sampling that can incur substantial cost because of numerous evaluations of expensive-to-evaluate high fidelity models as seen in Figure~\ref{fig:reliabilityAna} (a). There are several methods that improve the convergence rate of Monte Carlo methods to decrease computational cost through Monte Carlo variance reduction, such as, importance sampling~\cite{melchers1989importance,liu2008monte}, cross-entropy method~\cite{kroese2013cross}, subset simulation~\cite{au2001estimation,papaioannou2015mcmc}, etc. However, such methods are outside the scope of this paper and will not be discussed further. Another class of methods reduce the computational cost by using approximations for the failure boundary or the entire limit state function. The popular methods that fall in the first category are first- and second-order reliability methods (FORM and SORM), which approximate the failure boundary with linear and quadratic approximations around the most probable failure point~\cite{hohenbichler1987new,rackwitz2001reliability}. The FORM and SORM methods can be efficient for mildly nonlinear problems and cannot handle systems with multiple failure regions. The methods that fall in the second category reduce computational cost by replacing the high-fidelity model evaluations in the Monte Carlo simulation by cheaper evaluations from adaptive surrogates for the limit state function as seen in Figure~\ref{fig:reliabilityAna} (b).

Estimating reliability requires accurately classifying samples to fail or not, which needs surrogates that accurately predict the limit state function around the failure boundary. Thus, the surrogates need to be refined only in the region of interest (in this case, around the failure boundary) and do not require global accuracy in prediction of the limit state function. The development of sequential active learning methods for refining the surrogate around the failure boundary has been addressed in the literature using only a single high-fidelity information source. Such methods fall in the same category as adaptively refining surrogates for identifying stability boundaries, contour location, classification, sequential design of experiment (DOE) for target region, etc. Typically, these methods are divided into using either Gaussian process (GP) surrogates or support vector machines (SVM). Adaptive SVM methods have been implemeted for reliability analysis and contour location~\cite{basudhar2008limit,basudhar2013reliability,lecerf2015methodology}. In this work, we focus on GP-based methods (sometimes referred to as kriging-based) that use the GP prediction mean and prediction variance to develop greedy and lookahead adaptive sampling methods. Efficient Global Reliability Analysis (EGRA) adaptively refines the GP surrogate around the failure boundary by sequentially adding points that have maximum expected feasibility~\cite{bichon2008}. A weighted integrated mean square criterion for refining the kriging surrogate was developed by Picheny et al.~\cite{picheny2010adaptive}. Echard et al.~\cite{echard2011ak} proposed an adaptive Kriging method that refines the surrogate in the restricted set of samples defined by a Monte Carlo simulation. Dubourg et al.~\cite{dubourg2011reliability} proposed a population-based adaptive sampling technique for refining the kriging surrogate around the failure boundary. One-step lookahead strategies for GP surrogate refinement for estimating probability of failure was proposed by Bect et al.~\cite{bect2012sequential} and Chevalier et al.~\cite{chevalier2014fast}. A review of some surrogate-based methods for reliability analysis can be found in Ref.~\cite{moustapha2019surrogate}.
However, all the methods mentioned above use a single source of information, which is the high-fidelity model as illustrated in Figure~\ref{fig:reliabilityAna} (b). This work presents a novel multifidelity active learning method that adaptively refines the surrogate around the limit state function failure boundary using multiple sources of information, thus, further reducing the active learning computational effort as seen in Figure~\ref{fig:reliabilityAna} (c).

For several applications, in addition to an expensive high-fidelity model, there are potentially cheaper lower fidelity models, such as, simplified physics models, coarse-grid models, data-fit models, reduced order models, etc. that are readily available or can be built. This necessitates the development of multifidelity methods that can take advantage of these multiple information sources~\cite{peherstorfer2018survey}. Various multifidelity methods have been developed in the context of GP-based Bayesian optimization~\cite{poloczek2017multi,lam2015,ghoreishi2019multi}. While Bayesian optimization also uses GP models and adaptive sampling~\cite{frazier2018tutorial,jones2001taxonomy}, we note that Bayesian optimization targets a different problem to GP-based reliability analysis. In particular, the reliability analysis problem targets the entire limit state function failure contour in the random variable space, whereas Bayesian optimization targets finding the optimal design. Thus, the sampling criteria used for failure boundary location as compared to optimization are different, and the corresponding needs and opportunities for multifidelity methods are different.
In the context of reliability analysis using active learning surrogates, there are few multifidelity methods available. Dribusch et al.~\cite{dribusch2010multifidelity} proposed a hierarchical bi-fidelity adaptive SVM method for locating failure boundary. The recently developed CLoVER~\cite{marques2018contour} method is a multifidelity active learning algorithm that uses a one-step lookahead entropy-reduction-based adaptive sampling strategy for refining GP surrogates around the failure boundary. In this work, we develop a multifidelity extension of the EGRA method~\cite{bichon2008} as EGRA has been rigorously tested on a wide range of reliability analysis problems.

We propose \textbf{mfEGRA} (multifidelity EGRA) that leverages multiple sources of information with different fidelities and cost to accelerate active learning of surrogates for failure boundary identification. 
%The development of models for any application typically advance in stages, going from cheap simple models to expensive complex models [cite ben multifidelity], with every model providing some good information. Using information from multiple information sources with different fidelities and costs can further increase the efficiency of the adaptive sampling methods for locating the failure boundary. 
For single-fidelity methods, the adaptive sampling criterion chooses where to sample next to refine the surrogate around the failure boundary. The challenge in developing a multifidelity adaptive sampling criterion is that we now have to answer two questions -- \begin{enumerate*}[label=(\roman*)] \item where to sample next, and \item what information source to use for evaluating the next sample.\end{enumerate*} 
This work proposes a new adaptive sampling criterion that allows the use of multiple fidelity models. In our mfEGRA method, we combine the expected feasibility function used in EGRA with a proposed weighted lookahead information gain to define the adaptive sampling criterion for multifidelity case. We use the Kullback-Leibler (KL) divergence to quantify the information gain and derive a closed-form expression for the multifidelity GP case. The key advantage of the mfEGRA method is the reduction in computational cost compared to single-fidelity active learning methods because it can utilize additional information from multiple cheaper low-fidelity models along with the high-fidelity model information. We demonstrate the computational efficiency of the proposed mfEGRA using a multimodal analytic test problem and an acoustic horn problem with disjoint failure regions.

The rest of the paper is structured as follows. Section~\ref{s:2} provides the problem setup for reliability analysis using multiple information sources. Section~\ref{s:3} describes the details of the proposed mfEGRA method along with the complete algorithm. The effectiveness of mfEGRA is shown using an analytical multimodal test problem and an acoustic horn problem in Section~\ref{s:4}. The conclusions are presented in Section~\ref{s:5}.

% % % % % % % % % % % % % % % % % % % % % % % % % % % % % % % %
\section{Problem Setup}\label{s:2}
The inputs to the system are the $N_z$ random variables $Z\in\Omega \subseteq \mathbb{R}^{N_z}$ with the probability density function $\pi$, where $\Omega$ denotes the random sample space. The vector of a realization of the random variables $Z$ is denoted by $\bz$. 

The probability of failure of the system is $p_\tF = \mathbb{P}(g(Z)>0)$, where $g:\Omega \mapsto \mathbb{R}$ is the limit state function. In this work, without loss of generality, the failure of the system defined as $g(\bz)>0$. The failure boundary is defined as the zero contour of the limit state function, $g(\bz)=0$, and any other failure boundary, $g(\bz)=c$, can be reformulated as a zero contour (i.e., $g(\bz)-c=0$).

One way to estimate the probability of failure for nonlinear systems is Monte Carlo simulation. The Monte Carlo estimate of the probability of failure $\hat{p}_\tF$ is
\begin{equation}
\hat{p}_\tF=\frac{1}{m}\sum_{i=1}^{m}\mathbb{I}_{\failset}(\bz_i),
\end{equation}
where $\bz_i, i=1,\dots,m$ are $m$ samples from probability density $\pi$, $\failset = \{ \bz \ | \ \bz\in\Omega, g(\bz) > 0  \}$ is the failure set, and $\mathbb{I}_{\failset}:\Omega \mapsto \{0,1\}$ is the indicator function defined as
\begin{equation}
\label{e:indicator}
\mathbb{I}_{\failset}(\bz) = \left\{\begin{array}{ll}
1, & \bz \in \failset \\
0, & \text{else.}
\end{array}\right.
\end{equation}
The probability of failure estimation requires many evaluations of the expensive-to-evaluate high-fidelity model for the limit state function $g$, which can make reliability analysis computationally prohibitive. The computational cost can be substantially reduced by replacing the high-fidelity model evaluations with cheap-to-evaluate surrogate model evaluations. However, to make accurate estimations of $\hat{p}_\tF$ using a surrogate model, the zero-contour of the surrogate model needs to approximate the failure boundary well. Adaptively refining the surrogate around the failure boundary, while trading-off global accuracy, is an efficient way of addressing the above.

The goal of this work is to make the adaptive refinement of surrogate models around the failure boundary more efficient by using multiple models with different fidelities and costs instead of only using the high-fidelity model. We develop a multifidelity active learning method that utilizes multiple information sources to efficiently refine the surrogate to accurately locate the failure boundary.
Let $g_l:\Omega \mapsto \mathbb{R}, l\in\{0,\dots,k\}$ be a collection of $k+1$ models for $g$ with associated cost $c_l(\bz)$ at location $\bz$, where the subscript $l$ denotes the information source. We define the model $g_0$ to be the high-fidelity model for the limit state function. The $k$ low-fidelity models of $g$ are denoted by $l=1,\dots,k$.
%The observations from $g_l$ may be noisy and noisy observations are defined by the normal distribution $\mathcal{N}(g_l(\bz),\lambda_l(\bz))$, where $\lambda_l(\bz)$ is the known variance of the model $l$. 
We use a multifidelity surrogate to simultaneously approximate all information sources while encoding the correlations between them. The adaptively refined multifidelity surrogate model predictions are used for the probability of failure estimation. The Monte Carlo estimate of the probability of failure is then estimated using the refined multifidelity surrogate and is denoted here by $\hat{p}_\tF^\text{MF}$. Next we describe the multifidelity surrogate model used in this work and the multifidelity active learning method used to sequentially refine the surrogate around the failure boundary.

% % % % % % % % % % % % % % % % % % % % % % % % % % % % % % % % % % % % % % %
\section{mfEGRA: Multifidelity EGRA with Information Gain}\label{s:3}
In this section, we introduce multifidelity EGRA (mfEGRA) that leverages the $k+1$ information sources to efficiently build an adaptively refined multifidelity surrogate to locate the failure boundary.

%%%%%%%%%%%%%%%%%%%%%%%%%%%%%%%%%%%%%%%%%%%%%%%%%%%%%%%%%
\subsection{mfEGRA method overview}\label{s:3overview}
\begin{figure}[h]
	\centering
	\begin{tikzpicture}[scale=1, auto]
	\tikzstyle{decision} = [diamond, minimum width=1cm, minimum height=0.5cm, draw=black, fill=orange!30, text width=8em, text centered, inner sep=0pt]
	\tikzstyle{block} = [rectangle, draw, fill=blue!15, text width=12em, text centered, rounded corners, minimum height=0.5em, inner ysep=4pt]
	\tikzstyle{line} = [draw, very thick, color=black!50, -latex']
	\tikzstyle{cloud} = [draw, rectangle=dashed, fill=green!20, text centered, text width=7em, minimum height=0.5em]
	\tikzstyle{start} = [ellipse, draw=black, fill=red!20, text width=3em, text centered, minimum height=0.5em]
	\tikzstyle{container} = [draw, very thick, black!70, rectangle, inner xsep=0.8cm, inner ysep=0.28cm, rounded corners, fill=gray!15, minimum height=4cm]
	%
	% Place nodes
	\node [block] (DOE) {Get initial DOE and evaluate models};
	\node [cloud, right=1cm of DOE] (models) {$k+1$ models $g_l, l=0,\dots,k$};
	\node [block, below=0.4cm of DOE] (GP) {Build initial\\ multifidelity GP};
	\node [decision, below=0.6cm of GP] (while) {Is mfEGRA stopping criterion\\ met?};
	% If yes
	\node [block, left=1.5cm of while, text width=10em] (PoF) {Estimate probability of failure using the adaptively refined multifidelity GP};
	\node [start, below=0.4cm of PoF] (stop) {Stop};
	% If no
	\node [block, below=0.6cm of while] (location) {Select next sampling location using expected feasibility function};
	\node [block, below=0.4cm of location] (IS) {Select the information source using weighted lookahead information gain};
	\node [block, below=0.4cm of IS] (evaluate) {Evaluate at the next sample using the selected model};
	\node [block, below=0.4cm of evaluate] (update) {Update\\ multifidelity GP};
	% Draw box enclosing mfEGRA method
	\begin{scope}[on background layer]
	\node [container, label={[font=\large, black!70]right:\rotatebox{-90}{\textbf{mfEGRA}}}, fit=(while) (location) (IS) (evaluate) (update)] {};
	\end{scope}
	%
	% Draw edges
	\path [line] (DOE) -- (GP);
	\path [line] (GP) -- (while);
	\path [line] (while) -- node [near start, color=black] {Yes} (PoF);
	\path [line] (PoF) -- (stop);
	\path [line] (while) -- node [near start, color=black] {No}(location);
	\path [line] (location) -- (IS);
	\path [line] (IS) -- (evaluate);
	\path [line] (evaluate) -- (update);
	\path [line] (update) |- + (2.35,0) |- (while.east);
	\path [line,dashed] (models) -- (DOE);
	\path [line,dashed] (models) |- (evaluate);
	\end{tikzpicture}
	\caption{Flowchart showing the mfEGRA method.}
	\label{fig:mfEGRAflow}
\end{figure}
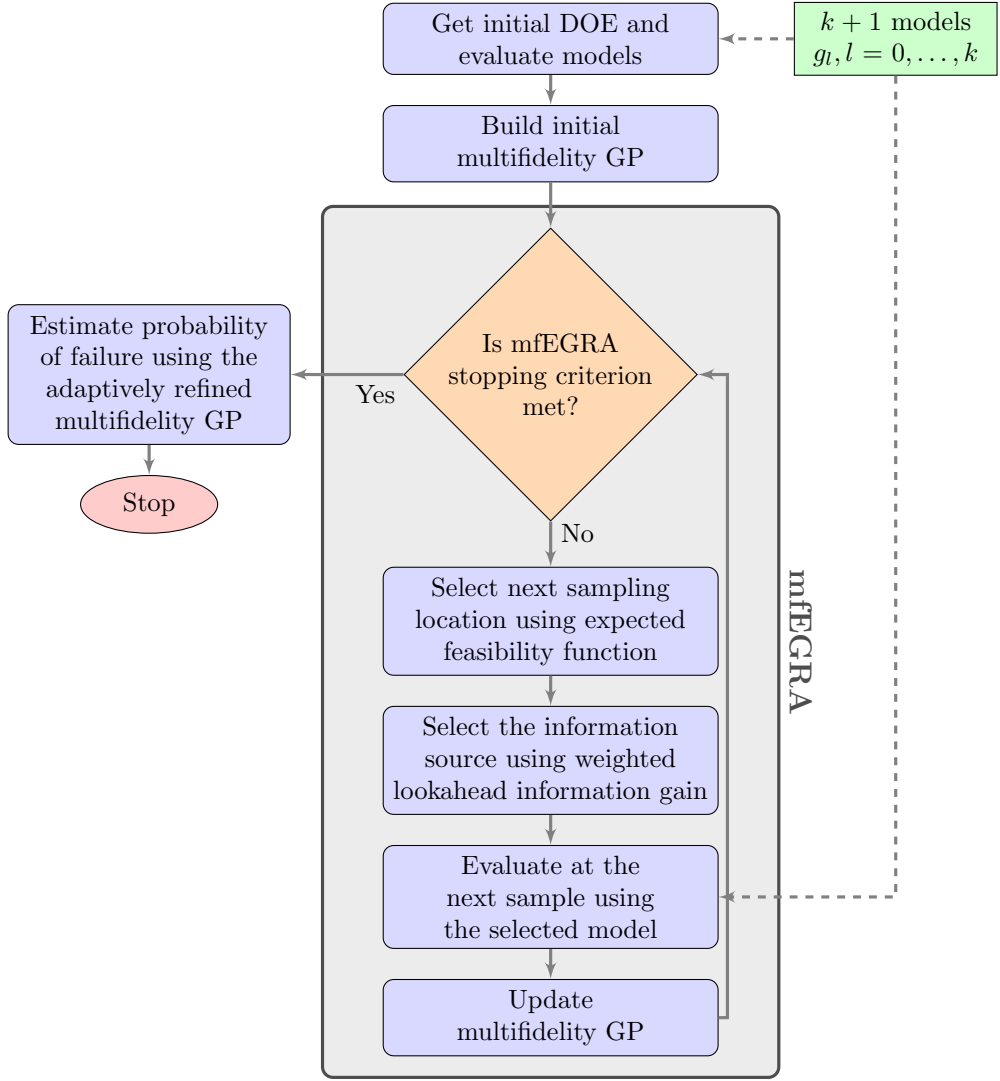

The proposed mfEGRA method is a multifidelity extension to the EGRA method~\cite{bichon2008}. Section~\ref{s:3a} briefly describes the multifidelity GP surrogate used in this work to combine the different information sources. The multifidelity GP surrogate is built using an initial DOE and then the mfEGRA method refines the surrogate using a two-stage adaptive sampling criterion that:
\begin{enumerate}
	\item selects the next location to be sampled using an expected feasibility function as described in Section~\ref{s:3b};
	\item selects the information source to be used to evaluate the next sample using a weighted lookahead information gain criterion as described in Section~\ref{s:3c}.	
\end{enumerate}
The adaptive sampling criterion developed in this work enables us to use the surrogate prediction mean and the surrogate prediction variance to make the decision of where and which information source to sample next. Note that both of these quantities are available from the multifidelity GP surrogate used in this work. Section~\ref{s:3d} provides the implementation details and the algorithm for the proposed mfEGRA method. Figure~\ref{fig:mfEGRAflow} shows a flowchart outlining the mfEGRA method.

%%%%%%%%%%%%%%%%%%%%%%%%%%%%%%%%%%%%%%%%%%%%%%%%%%%%%%%%%
\subsection{Multifidelity Gaussian process}\label{s:3a}
We use the multifidelity GP surrogate introduced by Poloczek et al.~ \cite{poloczek2017multi}, which built on earlier work by Lam et al.~\cite{lam2015}, to combine information from the $k+1$ information sources into a single GP surrogate, $\hg(l,\bz)$, that can simultaneously approximate all the information sources. The multifidelity GP surrogate can provide predictions for any information source $l$ and random variable realization $\bz$.

The multifidelity GP is built by making two modeling choices: (1) a GP approximation for the high-fidelity model $g_0$ as given by $\hg(0,\bz) \sim \GP(\mu_0,\Sigma_0)$, and (2) independent GP approximations for the model discrepancy between the high-fidelity and the lower-fidelity models as given by $\delta_l \sim \GP(\mu_l,\Sigma_l)$ for $l=1,\dots,k$. $\mu_l$ denotes the mean function and $\Sigma_l$ denotes the covariance kernel for $l=0,\dots,k$.

Then the surrogate for model $l$ is constructed by using the definition $\hg(l,\bz) = \hg(0,\bz)+\delta_l(\bz)$. These modeling choices lead to the surrogate model $\hg \sim \GP(\mu_\pr,\Sigma_\pr)$ with prior mean function $\mu_\pr$ and prior covariance kernel $\Sigma_\pr$. The priors for $l=0$ are
\begin{equation}
\label{e:muGP}
\begin{split}
\mu_\pr(0,\bz) &= \mathbb{E}[\hg(0,\bz)] = \mu_0(\bz), \\
\Sigma_\pr((0,\bz),(l',\bz')) &= \operatorname{Cov}\left(\hg(0,\bz),\hg(0,\bz')\right) = \Sigma_0(\bz,\bz'),
\end{split}
\end{equation}
and priors for $l=1,\dots,k$ are
\begin{equation}
\label{e:SigmaGP}
\begin{split}
\mu_\pr(l,\bz)= \mathbb{E}[\hg(l,\bz)] &= \mathbb{E}[\hg(0,\bz)] + \mathbb{E}[\delta_l(\bz)] = \mu_0(\bz) + \mu_l(\bz),\\
\Sigma_\pr((l,\bz),(l',\bz')) &= \operatorname{Cov}\left(\hg(0,\bz)+\delta_l(\bz),\hg(0,\bz')+\delta_{l'}(\bz')\right) \\
&= \Sigma_0(\bz,\bz') + \mathbbm{1}_{l,l'}\Sigma_l(\bz,\bz'),
\end{split}
%\begin{split}
%\Sigma_\pr((l,\bz),(l',\bz')) &= \operatorname{Cov}\left(\hg(0,\bz)+\delta_l(\bz),\hg(0,\bz')+\delta_{l'}(\bz')\right) \\
%&= \Sigma_0(\bz,\bz') + \mathbbm{1}_{l,l'}\Sigma_l(\bz,\bz'),
%\end{split}
\end{equation}
where $l'\in {0,\dots,k}$ and $\mathbbm{1}_{l,l'}$ denotes the Kronecker delta. Once the prior mean function and the prior covariance kernels are defined using Equations~\eqref{e:muGP}  and \eqref{e:SigmaGP}, we can compute the posterior using standard rules of GP regression~\cite{rasmussen2010gaussian}. A more detailed description about the assumptions and the implementation of the multifidelity GP surrogate can be found in Ref.~\cite{poloczek2017multi}.

At any given $\bz$, the surrogate model posterior distribution of $\hg(l,\bz)$ is defined by the normal distribution with posterior mean $\mu(l,\bz)$ and posterior variance $\sigma^2(l,\bz) = \Sigma((l,\bz),(l,\bz))$. Consider that $n$ samples $\{[l^i,\bz^i]\}_{i=1}^n$ have been evaluated and these samples are used to fit the present multifidelity GP surrogate. 
Note that $[l,\bz]$ is the augmented vector of inputs to the multifidelity GP. Then the surrogate is refined around the failure boundary by sequentially adding samples. The next sample $\bz^{n+1}$ and the next information source $l^{n+1}$ used to refine the surrogate are found using the two-stage adaptive sampling method mfEGRA as described below.

%%%%%%%%%%%%%%%%%%%%%%%%%%%%%%%%%%%%%%%%%%%%%%%%%%%%%%%%%
\subsection{Location selection: Maximize expected feasibility function}\label{s:3b}
The first stage of mfEGRA involves selecting the next location $\bz^{n+1}$ to be sampled. The expected feasibility function (EFF), which was used as the adaptive sampling criterion in EGRA~\cite{bichon2008}, is used in this work to select the location of the next sample $\bz^{n+1}$. The EFF defines the expectation of the sample lying within a band around the failure boundary (here, $\pm\epsilon(\bz)$ around the zero contour of the limit state function). The prediction mean $\mu(0,\bz)$ and the prediction variance $\sigma(0,\bz)$ at any $\bz$ are provided by the multifidelity GP for the high-fidelity surrogate model. The multifidelity GP surrogate prediction at $\bz$ is the normal distribution $\mathcal{Y}_{\bz} \sim \mathcal{N}(\mu(0,\bz),\sigma^2(0,\bz))$. Then the feasibility function at any $\bz$ is defined as being positive within the $\epsilon$-band around the failure boundary and zero otherwise as given by
\begin{equation}
	\label{e:Feas}
	F(\bz) = \epsilon(\bz) - \min (|y|,\epsilon(\bz)),
\end{equation}
where $y$ is a realization of $\mathcal{Y}_{\bz}$. The EFF is defined as the expectation of being within the $\epsilon$-band around the failure boundary as given by
\begin{equation}
	\label{e:EF_int}
	%\mathbb{E}_{\mathcal{G}_{\bz}}[F(\bz)] = \int_{-\infty}^{\infty} F(\bz)\mathcal{G}_{\bz}(\hg)\mathrm{d}\hg = \int_{-\epsilon}^{\epsilon} (\epsilon - |\hg|)\mathcal{G}_{\bz}(\hg)\mathrm{d}\hg.
	\mathbb{E}_{\mathcal{Y}_{\bz}}[F(\bz)] = \int_{-\epsilon(\bz)}^{\epsilon(\bz)} (\epsilon(\bz) - |y|)\mathcal{Y}_{\bz}(y)\mathrm{d}y.
\end{equation}
We will use $\mathbb{E}[F(\bz)]$ to denote $\mathbb{E}_{\mathcal{Y}_{\bz}}[F(\bz)]$ in the rest of the paper. The integration in Equation~\eqref{e:EF_int} can be solved analytically to obtain~\cite{bichon2008} 
\begin{equation}
	\label{e:EF}
	\begin{split}
		\mathbb{E}[F(\bz)] &= \mu(0,\bz) \left[2\Phi \left(\frac{-\mu(0,\bz)}{\sigma(0,\bz)}\right) - \Phi \left(\frac{-\epsilon(\bz)-\mu(0,\bz)}{\sigma(0,\bz)}\right) - \Phi \left(\frac{\epsilon(\bz)-\mu(0,\bz)}{\sigma(0,\bz)}\right) \right] \\
		& - \sigma(0,\bz) \left[2\phi \left(\frac{-\mu(0,\bz)}{\sigma(0,\bz)}\right) - \phi\left(\frac{-\epsilon(\bz)-\mu(0,\bz)}{\sigma(0,\bz)}\right) - \phi\left(\frac{\epsilon(\bz)-\mu(0,\bz)}{\sigma(0,\bz)}\right) \right] \\
		& + \epsilon(\bz) \left[\Phi \left(\frac{\epsilon(\bz)-\mu(0,\bz)}{\sigma(0,\bz)}\right) - \Phi \left(\frac{-\epsilon(\bz)-\mu(0,\bz)}{\sigma(0,\bz)}\right) \right],
	\end{split}
\end{equation} 
where $\Phi$ is the cumulative distribution function and $\phi$ is the probability density function of the standard normal distribution. Similar to EGRA~\cite{bichon2008}, we define $\epsilon(\bz)=2\sigma(0,\bz)$ to balance exploration and exploitation. As noted before, we describe the method considering the zero contour as the failure boundary for convenience but the proposed method can be used for locating failure boundary at any contour level.

The location of the next sample is selected by maximizing the EFF as given by
\begin{equation}
	\label{e:Loc}
	\bz^{n+1}=\argmax_{\bz\in\Omega} \mathbb{E}[F(\bz)].
\end{equation}

%%%%%%%%%%%%%%%%%%%%%%%%%%%%%%%%%%%%%%%%%%%%%%%%%%%%%%%%%%%%%%%%%%%%%%%%
\subsection{Information source selection: Maximize weighted lookahead information gain}\label{s:3c}
Given the location of the next sample at $\bz^{n+1}$ obtained using Equation~\eqref{e:Loc}, the second stage of mfEGRA selects the information source $l^{n+1}$ to be used for simulating the next sample by maximizing the information gain. Information-gain-based approaches have been used previously for global optimization~\cite{villemonteix2009,hennig2012,ghoreishi2019multi,hernandez2014}, optimal experimental design~\cite{huan2013,villanueva2015}, and uncertainty propagation in coupled systems~\cite{chaudhuri2018multifidelity}. Ref.~\cite{ghoreishi2019multi} used an information-gain-based approach for selecting the location and the information source for improving the global accuracy of the multifidelity GP approximation for the constraints in global optimization using a double-loop Monte Carlo sample estimate of the information gain. Our work differs from previous efforts in that we develop a weighted information-gain-based sampling strategy for failure boundary identification utilizing multiple fidelity models. In the context of information gain criterion for multifidelity GP, the two specific contributions of our work are: (1) deriving a closed-form expression for KL divergence for the multifidelity GP, which does not require double-loop Monte Carlo sampling, thus improving the robustness and decreasing the cost of estimation of the acquisition function, and (2) using weighting strategies to address the goal of failure boundary location for reliability analysis using multiple fidelity models.

The next information source is selected by using a weighted one-step lookahead information gain criterion. This adaptive sampling strategy selects the information source that maximizes the information gain in the GP surrogate prediction defined by the Gaussian distribution at any $\bz$.
% In this work, information gain is quantified by the KL divergence. 
We measure the KL divergence between the present surrogate predicted GP and a hypothetical future surrogate predicted GP when a particular information source is used to simulate the sample at $\bz^{n+1}$ to quantify the information gain.

We represent the present GP surrogate built using the $n$ available training samples by the subscript $\tP$ for convenience as given by $\hg_\tP(l,\bz)=\hg(l,\bz\ |\ \{l^i,\bz^i\}_{i=1}^n)$. Then the present surrogate predicted Gaussian distribution at any $\bz$ is 
$$
G_\tP(\bz)\sim \mathcal{N}(\mu_\tP(0,\bz),\sigma_\tP^2(0,\bz)),
$$ 
where $\mu_\tP(0,\bz)$ is the posterior mean and $\sigma_\tP^2(0,\bz)$ is the posterior prediction variance of the present GP surrogate for the high-fidelity model built using the available training data till iteration $n$. 

A hypothetical future GP surrogate can be understood as a surrogate built using the current GP as a generative model to create hypothetical future simulated data. The hypothetical future simulated data $y_\tF \sim \mathcal{N}(\mu_\tP(l_\tF,\bz^{n+1}), \sigma_\tP^2(l_\tF,\bz^{n+1}))$ is obtained from the present GP surrogate prediction at the location $\bz^{n+1}$ using a possible future information source $l_F \in \{0,\dots,k\}$.  We represent a hypothetical future GP surrogate by the subscript $\tF$. 
Then a hypothetical future surrogate predicted Gaussian distribution at any $\bz$ is
$$
G_\tF(\bz|\bz^{n+1},l_\tF, y_\tF)\sim \mathcal{N}(\mu_\tF(0,\bz|\bz^{n+1},l_\tF, y_\tF),\sigma_\tF^2(0,\bz|\bz^{n+1},l_\tF, y_\tF)).
$$
The posterior mean of the hypothetical future GP is affine with respect to $y_\tF$ and thus is distributed normally as given by
$$
\mu_\tF(0,\bz|\bz^{n+1},l_\tF, y_\tF) \sim \mathcal{N}(\mu_\tP(0,\bz),\bar{\sigma}^2(\bz|\bz^{n+1},l_\tF)),
$$
where 
%$\bar{\sigma}^2(\bz|\bz^{n+1},l_\tF) = (\Sigma_\tP((0,\bz),(l_\tF,\bz^{n+1})))^2/(\lambda_l(\bz) + \Sigma_\tP((l_\tF,\bz^{n+1}),(l_\tF,\bz^{n+1})))$ 
$\bar{\sigma}^2(\bz|\bz^{n+1},l_\tF) = (\Sigma_\tP((0,\bz),(l_\tF,\bz^{n+1})))^2/\Sigma_\tP((l_\tF,\bz^{n+1}),(l_\tF,\bz^{n+1}))$\cite{poloczek2017multi}.
The posterior variance of the hypothetical future GP surrogate $\sigma_\tF^2(0,\bz|\bz^{n+1},l_\tF, y_\tF)$ depends only on the location $\bz^{n+1}$ and the source $l_\tF$, and can be replaced with $\sigma_\tF^2(0,\bz|\bz^{n+1},l_\tF)$. Note that we don't need any new evaluations of the information source for constructing the future GP. The total lookahead information gain is obtained by integrating over all possible values of $y_\tF$ as described below.

Since both $G_\tP$ and $G_\tF$ are Gaussian distributions, we can write the KL divergence between them explicitly. The KL divergence between $G_\tP$ and $G_\tF$ for any $\bz$ is
\begin{equation}
	\label{e:KL}
	\begin{split}
		&D_{\text{KL}}(G_\tP(\bz)\parallel G_\tF(\bz|\bz^{n+1},l_\tF, y_\tF))\\
		&=\log\left(\frac{\sigma_\tF(0,\bz|\bz^{n+1},l_\tF)}{\sigma_\tP(0,\bz)}\right)+\frac{\sigma_\tP^2(0,\bz)+(\mu_\tP(0,\bz)-\mu_\tF(0,\bz|\bz^{n+1},l_\tF, y_\tF))^2}{2\sigma_\tF^2(0,\bz|\bz^{n+1},l_\tF)}-\frac{1}{2}.
	\end{split}
\end{equation}
The total KL divergence can then be calculated by integrating $D_{\text{KL}}(G_\tP(\bz)\parallel G_\tF(\bz|\bz^{n+1},l_\tF, y_\tF))$ over the entire random variable space $\Omega$ as given by
\begin{equation}
	\label{e:KLtot}
	\begin{split}
		&\int_{\Omega} D_{\text{KL}}(G_\tP(\bz)\parallel G_\tF(\bz|\bz^{n+1},l_\tF, y_\tF)) \mathrm{d}z\\
		&=\int_{\Omega} \left[\log\left(\frac{\sigma_\tF(0,\bz|\bz^{n+1},l_\tF)}{\sigma_\tP(0,\bz)}\right)+\frac{\sigma_\tP^2(0,\bz)+(\mu_\tP(0,\bz)-\mu_\tF(0,\bz|\bz^{n+1},l_\tF, y_\tF))^2}{2\sigma_\tF^2(0,\bz|\bz^{n+1},l_\tF)}-\frac{1}{2}\right] \mathrm{d}\bz.
	\end{split}
\end{equation}
The total lookahead information gain for any $\bz$ can then be calculated by taking the expectation of Equation~\eqref{e:KLtot} over all possible values of $y_\tF$ as given by
\begin{equation}
	\label{e:totalIG}
	\begin{split}
		&D_{\IG}(\bz^{n+1},l_\tF) = \mathbb{E}_{y_\tF}\left[\int_{\Omega} D_{\text{KL}}(G_\tP(\bz)\parallel G_\tF(\bz|\bz^{n+1},l_\tF, y_\tF)) \mathrm{d}\bz \right] \\
		& = \int_{\Omega} \left[\log\left(\frac{\sigma_\tF(0,\bz|\bz^{n+1},l_\tF)}{\sigma_\tP(0,\bz)}\right)+\frac{\sigma_\tP^2(0,\bz)+\mathbb{E}_{y_\tF}\left[(\mu_\tP(0,\bz)-\mu_\tF(0,\bz|\bz^{n+1},l_\tF, y_\tF))^2\right]}{2\sigma_\tF^2(0,\bz|\bz^{n+1},l_\tF)}-\frac{1}{2} \right]\mathrm{d}\bz \\
		& = \int_{\Omega} \left[\log\left(\frac{\sigma_\tF(0,\bz|\bz^{n+1},l_\tF)}{\sigma_\tP(0,\bz)}\right)+\frac{\sigma_\tP^2(0,\bz)+\bar{\sigma}^2(\bz|\bz^{n+1},l_\tF)}{2\sigma_\tF^2(0,\bz|\bz^{n+1},l_\tF)}-\frac{1}{2} \right]\mathrm{d}\bz \\
		& = \int_{\Omega}D(\bz\ |\ \bz^{n+1},l_\tF)\mathrm{d}\bz,
	\end{split}
\end{equation}
where $$D(\bz\ |\ \bz^{n+1},l_\tF) = \log\left(\frac{\sigma_\tF(0,\bz|\bz^{n+1},l_\tF)}{\sigma_\tP(0,\bz)}\right)+\frac{\sigma_\tP^2(0,\bz)+\bar{\sigma}^2(\bz|\bz^{n+1},l_\tF)}{2\sigma_\tF^2(0,\bz|\bz^{n+1},l_\tF)}-\frac{1}{2}. $$
In practice, we choose a discrete set $\mathcal{Z}\subset\Omega$ via Monte Carlo sampling to numerically integrate Equation~\eqref{e:totalIG} as given by
\begin{equation}
	\label{e:DIG_num}
	D_{\IG}(\bz^{n+1},l_\tF) = \int_{\Omega}D(\bz\ |\ \bz^{n+1},l_\tF)\mathrm{d}\bz \propto \sum_{\bz \in \mathcal{Z}}D(\bz\ |\ \bz^{n+1},l_\tF).
\end{equation}
The total information gain for the multifidelity GP can be estimated using single-loop Monte Carlo sampling instead of double-loop Monte Carlo sampling because of the closed-form expression derived in Equation~\eqref{e:totalIG}. This improves the robustness and decreases the cost of estimation of the acquisition function.

The total lookahead information gain evaluated using Equation~\eqref{e:DIG_num} gives a metric of global information gain over the entire random variable space. However, we are interested in gaining more information around the failure boundary. In order to give more importance to gaining information around the failure boundary we use a weighted version of the lookahead information gain normalized by the cost of the information source.
In this work, we explore three different weighting strategies: (i) no weights $w(\bz)=1$, (ii) weights defined by the EFF, $w(\bz)=\mathbb{E}[F(\bz)]$, and (iii) weights defined by the probability of feasibility (PF), $w(\bz)=\mathbb{P}[F(\bz)]$. The PF of the sample to lie within the $\pm\epsilon(\bz)$ bounds around the zero contour is
\begin{equation}
	\mathbb{P}[F(\bz)] = \Phi\left( \frac{\epsilon(\bz)-\mu(0,\bz)}{\sigma(0,\bz)} \right) - \Phi\left( \frac{-\epsilon(\bz)-\mu(0,\bz)}{\sigma(0,\bz)} \right).
\end{equation} 
Weighting the information gain by either expected feasibility or probability of feasibility gives more importance to gaining information around the target region, in this case, the failure boundary. 

The next information source $l^{n+1}$ is selected by maximizing the weighted lookahead information gain normalized by the cost of the information source as given by
\begin{equation}
	\label{e:IS}
	l^{n+1}=\argmax_{l\in\{0,\dots,k\}} \sum_{\bz \in \mathcal{Z}} \frac{1}{c_l(\bz)} w(\bz)D(\bz|\bz^{n+1},l_\tF=l).
\end{equation}
Note that the optimization problem in Equation~\eqref{e:IS} is a one-dimensional discrete variable problem. In this case, we only need $k+1$ (number of available models) evaluations of the objective function to solve the optimization problem exactly and typically $k$ is a small number.

%%%%%%%%%%%%%%%%%%%%%%%%%%%%%%%%%%%%%%%%%%%%%%%%%%%%%%%%
\subsection{Algorithm and implementation details}\label{s:3d}
An algorithm describing the mfEGRA method is given in Algorithm~\ref{a:mfEGRA}. In this work, we evaluate all the models at the initial DOE. We generate the initial samples ${\bz}$ using Latin hypercube sampling and run all the models at each of those samples to get the initial training set $\{\bz^i,l^i\}_{i=1}^n$. The initial number of samples, $n$, can be decided based on the user's preference (in this work, we use cross-validation error). The EFF maximization problem given by Equation~\eqref{e:Loc} is solved using \texttt{patternsearch} function followed by using multiple starts of a local optimizer through the \texttt{GlobalSearch} function in MATLAB. In practice, we choose a fixed set of realizations $\mathcal{Z}\in\Omega$ at which the information gain is evaluated as shown in Equation~\eqref{e:DIG_num} for all iterations of mfEGRA. Due to the typically high cost associated with the high-fidelity model, we chose to evaluate all the $k+1$ models when the high-fidelity model is selected as the information source and update the GP hyperparameters in our implementation. All the $k+1$ model evaluations can be done in parallel. The algorithm is stopped when the maximum value of EFF goes below $10^{-10}$. However, other stopping criteria can also be explored.

Although in this work we did not encounter any case of failed model evaluations, numerical solvers can sometimes fail to provide a converged result. In the context of reliability analysis, a failed model evaluation can be treated as failure of the system (defined here as $g_l(\bz)>0$) at the particular random variable realization $\bz$. One possibility to handle failed model evaluations would be to let the value of the limit state function $g_l(\bz)$ go to an upper limit in order to indicate failure of the system.

A potential limitation of any GP-based method is dealing with the curse of dimensionality for high-dimensional problems, where the number of samples to cover the space grows exponentially and the cost of training GPs scales as the cube of the number of samples. The multifidelity method presented here alleviates the cost of exploring the space by using cheaper low-fidelity model evaluations and restricts its queries of the high-fidelity model to lie mostly around the failure boundary. The issue of the cost of training GPs with increasing number of samples is not addressed here but can be potentially tackled through GP sparsification techniques~\cite{williams2006gaussian,burt2019rates}. Another strategy for reducing cost of training is through adaptive sampling strategies that exploit parallel computing. Advancements in parallel computing have led to several parallel adaptive sampling strategies for global optimization~\cite{haftka2016parallel} and some parallel adaptive sampling methods for contour location~\cite{viana2012sequential,chevalier2014fast}. In addition, parallel methods for multifidelity adaptive sampling have increased difficulty and needs to be explored in both the fields of global optimization and contour location.

\begin{algorithm}[!htb]
	\caption{Multifidelity EGRA}
	\label{a:mfEGRA}
	\algorithmicrequire{ \ Initial DOE $\mathcal{X}_0=\{\bz^i,l^i\}_{i=1}^n$}, cost of each information source $c_l$ \\
	\algorithmicensure{ \ Refined multifidelity GP $\hg$}
	\begin{algorithmic}[1]
		\Procedure{mfEGRA}{$\mathcal{X}_0$}
		\State $\mathcal{X} = \mathcal{X}_0$	\Comment{set of training samples}
		\State Build initial multifidelity GP $\hg$ using the initial set of training samples $\mathcal{X}_0$
		%		\State Generate fixed set of realizations $\mathcal{Z}\subset\Omega$	\Comment{see Equation discretizing information gain estimation}
		\While{stopping criterion is not met}
		\State Select next sampling location $\bz^{n+1}$ using Equation~\eqref{e:Loc}
		\State Select next information source $l^{n+1}$ using Equation~\eqref{e:IS}
		\State Evaluate at sample $\bz^{n+1}$ using information source $l^{n+1}$
		\State $\mathcal{X} = \mathcal{X}\cup \{\bz^{n+1},l^{n+1}\}$
		\State Build updated multifidelity GP $\hg$ using $\mathcal{X}$
		\State $n \gets n+1$
		\EndWhile
		
		\State \Return $\hg$
		\EndProcedure
	\end{algorithmic}
\end{algorithm}

%%%%%%%%%%%%%%%%%%%%%%%%%%%%%%%%%%%%%%%%%%
\section{Results}\label{s:4}
In this section, we demonstrate the effectiveness of the proposed mfEGRA method on an analytic multimodal test problem and two different cases for an acoustic horn application. The probability of failure is estimated through Monte Carlo simulation using the adaptively refined multifidelity GP surrogate.

%%%%%%%%%%%%%%%%%%%%%%%%%%%%%%%%%%%%%%%%%%%%%%%%%%%%%%%%%%%%%%%
\subsection{Analytic multimodal test problem}\label{s:test}
\begin{figure}[!hbt]
	\centering
	\includegraphics[width=\textwidth, trim = {2.3cm 0cm 1cm 0cm},clip,page=1]{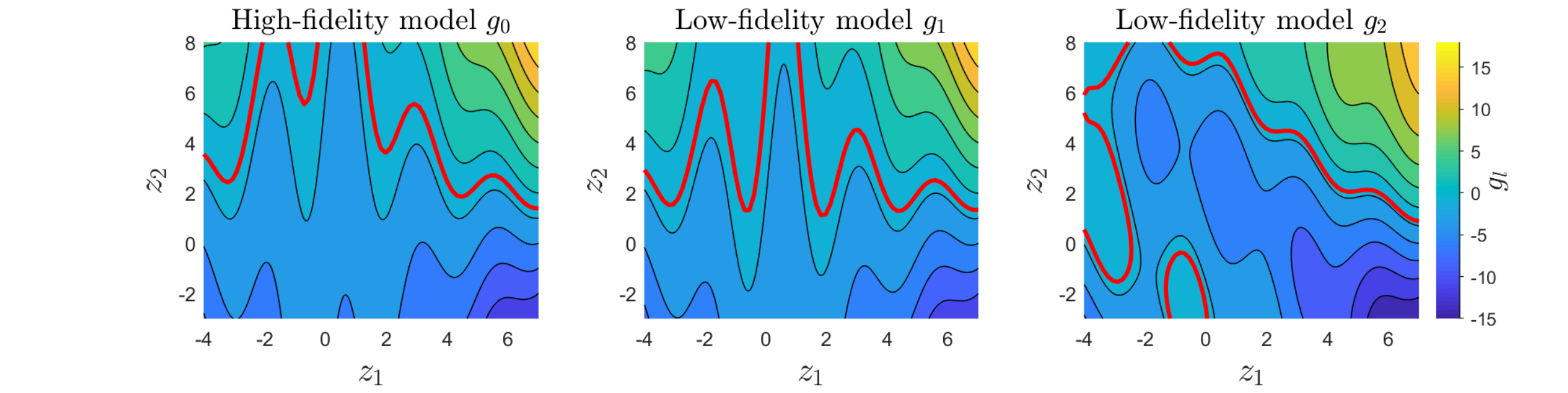}\hfill
	\caption{Contours of $g_l(\bz)$ using the three fidelity models for the analytic test problem. Solid red line represents the zero contour that denotes the failure boundary.}
	\label{fig:fidContour}
\end{figure}

The analytic test problem used in this work has two inputs and three models with different fidelities and costs. This test problem has been used before in the context of reliability analysis in Ref.~\cite{bichon2008}. The high-fidelity model of the limit state function is
\begin{equation}
	g_0(\bz) = \frac{(z_1^2+4)(z_2-1)}{20} - \sin\left(\frac{5z_1}{2}\right)-2,
\end{equation}
where $z_1\sim \mathcal{U}(-4,7)$ and $z_2\sim \mathcal{U}(-3,8)$ are uniformly distributed random numbers. The domain of the function is $\Omega = [-4, 7]\times[-3,8]$. The two low-fidelity models are
\begin{align}
	g_1(\bz) &= g_0(\bz) + \sin\left(\frac{5z_1}{22}  + \frac{5z_2}{44} + \frac{5}{4}\right), \\
	g_2(\bz) &= g_0(\bz) + 3\sin\left(\frac{5z_1}{11}  + \frac{5z_2}{11} + \frac{35}{11}\right).
\end{align}
The cost of each fidelity model is taken to be constant over the entire domain and is given by $c_0 = 1, c_1 = 0.01 \text{ and } c_2 = 0.001$. In this case, there is no noise in the observations from the different fidelity models. The failure boundary is defined by the zero contour of the limit state function ($g_0(\bz)=0$) and the failure of the system is defined by $g_0(\bz)>0$. Figure~\ref{fig:fidContour} shows the contour plot of $g(\bz)$ for the three models used for the analytic test problem along with the failure boundary for each of them.

We use an initial DOE of size 10 generated using Latin hypercube sampling. All the models are evaluated at these 10 samples to build the initial multifidelity surrogate. The reference probability of failure is estimated to be $\hat{p}_\tF=0.3021$ using $10^6$ Monte Carlo samples of $g_0$ model. The relative error in probability of failure estimate using the adaptively refined multifidelity GP surrogate, defined by $\lvert\hat{p}_\tF-\hat{p}_\tF^\text{MF}\rvert/\hat{p}_\tF$, is used to assess the accuracy and computational efficiency of the proposed method. We repeat the calculations for 100 different initial DOEs to get the confidence bands on the results.

\begin{figure}[!hbt]
	\centering
	\includegraphics[width=10cm, trim = {0cm 0cm 1cm 0.5cm},clip,page=2]{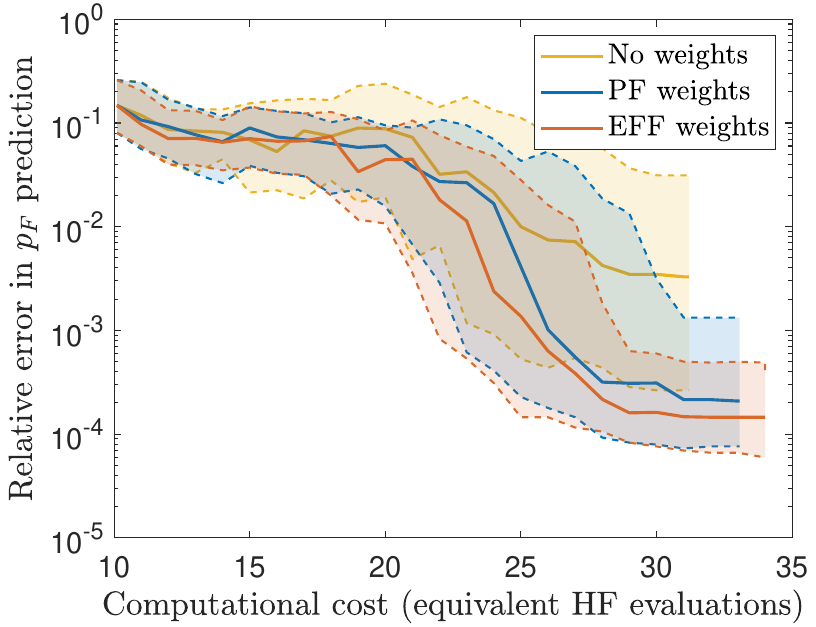}
	\caption{Effect of different weights for information gain criterion in mfEGRA for analytic test problem in terms of convergence of relative error in $p_\tF$ prediction (shown in log-scale) for 100 different initial DOEs. Solid lines represent the median and dashed lines represent the 25 and 75 percentiles.}
	\label{fig:diffWts}
\end{figure}

We first compare the accuracy of the method when different weights are used for the information gain criterion in mfEGRA as seen in Figure~\ref{fig:diffWts}. We can see that using weighted information gain (both EFF and PF) performs better than the case when no weights are used when comparing the error confidence bands. EFF-weighted information gain leads to only marginally lower errors in this case as compared to PF-weighted information gain. Since we don't see any significant advantage of using PF as weights and we use the EFF-based criterion to select the sample location, we propose using EFF-weighted information gain to make the implementation more convenient. Note that for other problems, it is possible that PF-weighted information gain may be better. From hereon, mfEGRA is used with the EFF-weighted information gain.

\begin{figure}[!hbt]
	\centering
	\includegraphics[width=10cm, trim = {0cm 0cm 1cm 0.5cm},clip,page=3]{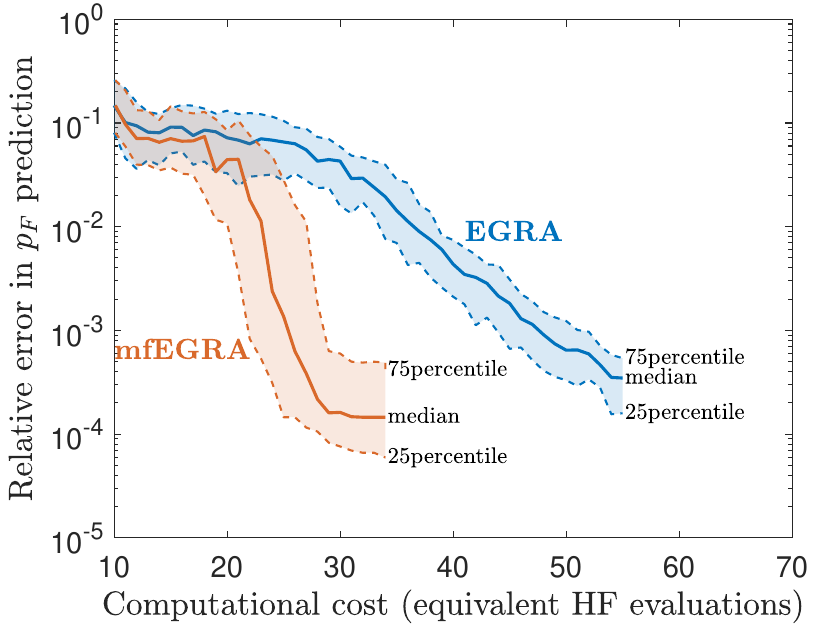}
	\caption{Comparison of mfEGRA vs single-fidelity EGRA for analytic test problem in terms of convergence of relative error in $p_\tF$ prediction (shown in log-scale) for 100 different initial DOEs.}
	\label{fig:mfEGRAvsEGRA}
\end{figure}

The comparison of mfEGRA with single-fidelity EGRA shows considerable improvement in accuracy at substantially lower computational cost as seen in Figure~\ref{fig:mfEGRAvsEGRA}. In this case, to reach a median relative error of below $10^{-3}$ in $p_\tF$ prediction, mfEGRA requires a computational cost of 26 compared to EGRA that requires a computational cost of 48 ($\sim$46\% reduction). Note that we start both cases with the same 100 sets of initial samples. We also note that the original paper for the EGRA method~\cite{bichon2008} reports a computational cost of 35.1 for the mean relative error from 20 different initial DOEs to reach below $5\times 10^{-3}$. We report the computational cost for the EGRA algorithm to reach a median relative error from 100 different initial DOEs below $10^{-3}$ to be 48 (in our case, the computational cost for the median relative error for EGRA to reach below $5\times 10^{-3}$ is 40). The difference in results can be attributed to the different sets of initial DOEs, the GP implementations, different statistics of reported results, and different probability distributions used for the random variables.

\begin{figure}[!hbt]
	\centering
	\subfigure[]{\includegraphics[width=0.5\textwidth, trim = {0cm 0cm 1cm 0.5cm},clip,page=4]{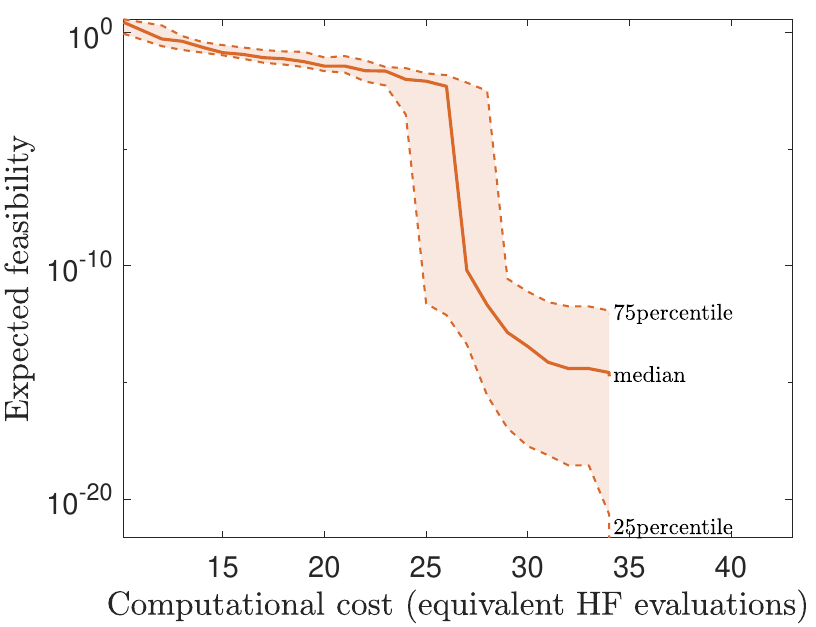}}\hfill
	\subfigure[]{\includegraphics[width=0.5\textwidth, trim = {0cm 0cm 1cm 0.5cm},clip,page=5]{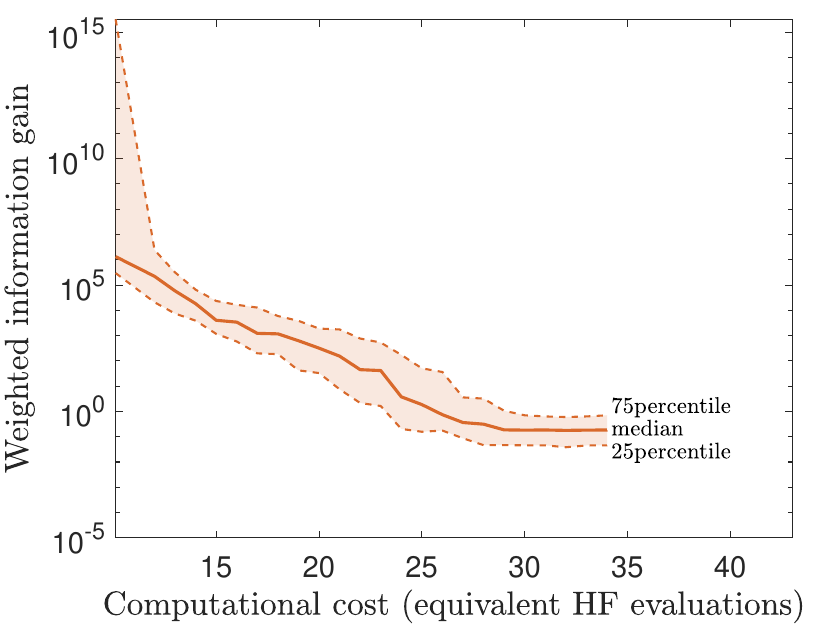}}
	\caption{Evolution of adaptive sampling criteria (a) expected feasibility function, and (b) weighted information gain used in mfEGRA for 100 different initial DOEs.}
	\label{fig:EF_IG}
\end{figure}

\begin{figure}[!]
	\centering
	\includegraphics[width=12cm, trim = {1.5cm 0cm 1.8cm 0.3cm},clip,page=6]{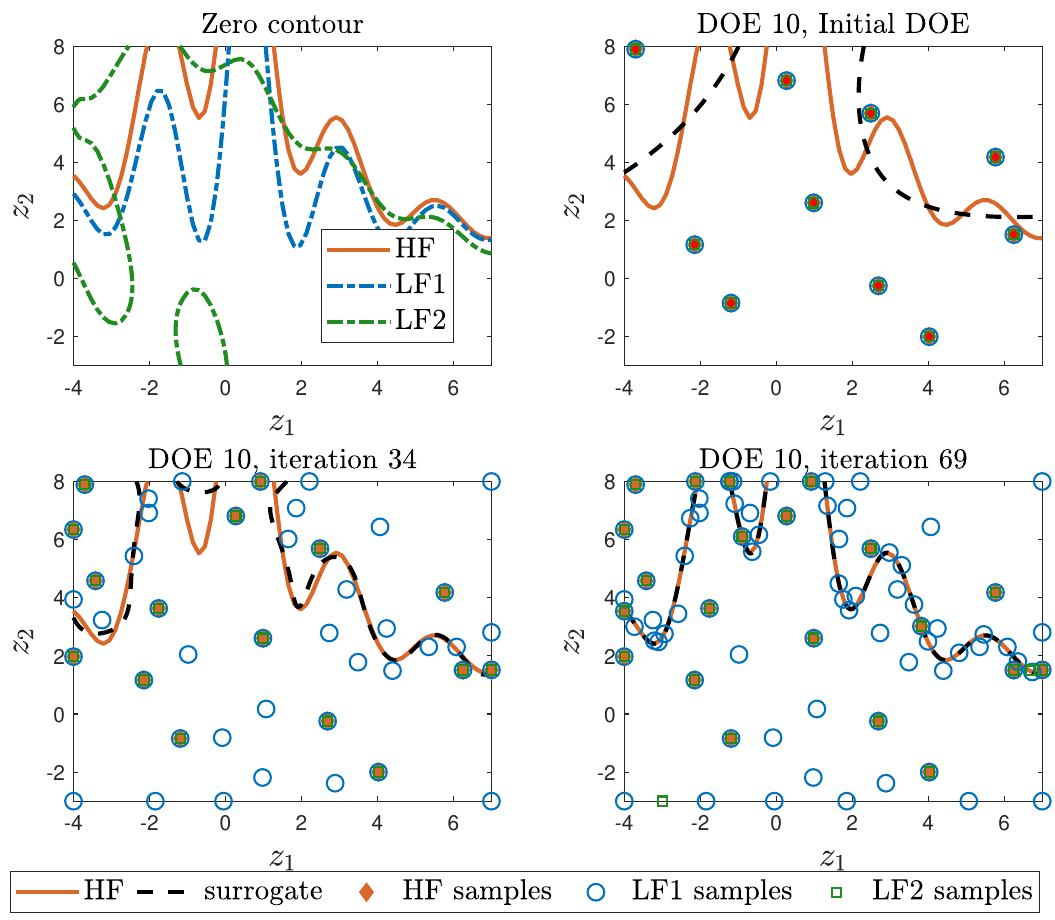}
	\caption{Progress of mfEGRA at several iterations showing the surrogate prediction and the samples from different models for a particular initial DOE. HF refers to high-fidelity model $g_0$, LF1 refers to low-fidelity model $g_1$, and LF2 refers to low-fidelity model $g_2$.}
	\label{fig:mfEGRAprog}
\end{figure}

Figure~\ref{fig:EF_IG} shows the evolution of the expected feasibility function and the weighted lookahead information gain, which are the two stages of the adaptive sampling criterion used in mfEGRA. These metrics along with the relative error in probability of failure estimate can be used to define an efficient stopping criterion, specifically when the adaptive sampling needs to be repeated for different sets of parameters (e.g., in reliability-based design optimization).
%A potentially interesting future work would be to analyze the metrics along with the relative error plot in Figure~\ref{fig:mfEGRAvsEGRA} to come up with a tuned application-specific stopping criteria instead of using a conservative stopping criterion used in this work. This could be useful for increasing efficiency when the adaptive sampling needs to be repeated for different sets of parameters (for e.g., in reliability-based design optimization).
%
%According to the desired level of relative error in the probability of failure estimate, the stopping criterion can be tuned by using different threshold values for EFF and weighted information gain.
%
Figure~\ref{fig:mfEGRAprog} shows the progress of mfEGRA at several iterations for a particular initial DOE. mfEGRA explores most of the domain using the cheaper $g_1$ and $g_2$ models in this case. The algorithm is stopped after 69 iterations when the expected feasibility function reached below $10^{-10}$; we can see that the surrogate contour accurately traces the true failure boundary defined by the high-fidelity model. As noted before, we evaluate all the three models when the high-fidelity model is selected as the information source. In this case, mfEGRA makes a total of 21 evaluations of $g_0$, 77 evaluations of $g_1$, and 23 evaluations of $g_2$ including the initial DOE, to reach a value of EFF below $10^{-10}$.

\subsection{Acoustic horn}\label{s:acoustic}
\begin{figure}[!hbt]
	\centering
	\includegraphics[width=8cm, trim = {4.6cm 16.5cm 6.6cm 2.5cm},clip,page=1]{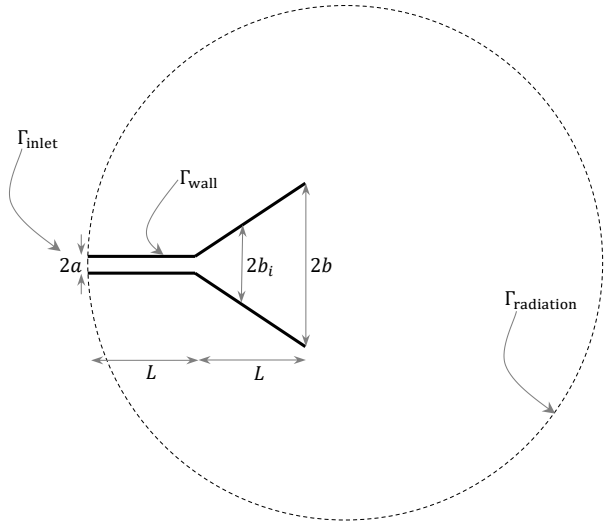}
	\caption{Two-dimensional acoustic horn geometry with $a = 0.5, b=3, L=5$ and shape of the horn flare described by six equally-spaced half-widths $b_1=0.8, b_2=1.2, b_3=1.6, b_4=2, b_5=2.3, b_6=2.65$.~\cite{ng2014multifidelity}}
	\label{fig:AchornIllus}
\end{figure}

We demonstrate the effectiveness of mfEGRA for the reliability analysis of an acoustic horn for a three-dimensional case and a rarer event probability of failure four-dimensional case. The acoustic horn model used in this work has been used in the context of robust optimization by Ng et al.~\cite{ng2014multifidelity} An illustration of the acoustic horn is shown in Figure~\ref{fig:AchornIllus}. 

\subsubsection{Three-dimensional case}
The inputs to the system are the three random variables listed in Table~\ref{t:randvar}.
\begin{table}[!htb]
	\centering
	\caption{Random variables used in the three-dimensional acoustic horn problem.}
	\label{t:randvar}
	\begin{tabular}{C{2cm}L{3cm}C{2cm}C{1cm}C{1cm}C{1cm}C{2cm}}
		\hline
		Random variable & Description & Distribution & Lower bound & Upper bound & Mean & Standard deviation \\ 
		\hline
		$k$ & wave number & Uniform & 1.3 & 1.5 & -- & --  \\
		$Z_u$ & upper horn wall impedance & Normal & -- & -- & 50 & 3 \\
		$Z_l$ & lower horn wall impedance & Normal & -- & -- & 50 & 3  \\
		\hline
	\end{tabular}
\end{table}

The output of the model is the reflection coefficient $s$, which is a measure of the horn's efficiency. We define the failure of the system to be $s(\bz)>0.1$. The limit state function is defined as $g(\bz) = s(\bz)-0.1$, which defines the failure boundary as $g(\bz)=0$. We use a two-dimensional acoustic horn model governed by the non-dimensional Helmholtz equation. In this case, a finite element model of the Helmholtz equation is the high-fidelity model $g_0$ with 35895 nodal grid points. The low-fidelity model $g_1$ is a reduced basis model with $N=100$ basis vectors~\cite{ng2014multifidelity,eftang2012two}. 
%The reduced basis model is obtained by projecting the system of equations from high-fidelity finite element model to a reduced subspace~\cite{eftang2012two}. 
%In this case, $g_1$ is a reduced basis model with $N=100$ reduced basis functions. 
%$c_0 = 0.3$ seconds and the low-fidelity model is $c_1 = 0.0075$ seconds,
In this case, the cost of evaluating the low-fidelity model is 40 times faster than evaluating the high-fidelity model. The cost of evaluating the different models is taken to be constant over the entire random variable space. A more detailed description of the acoustic horn models used in this work can be found in Ref.~\cite{ng2014multifidelity}.

\begin{figure}[!hbt]
	\centering
	\includegraphics[width=12cm, trim = {0cm 0cm 1cm 1cm},clip,page=1]{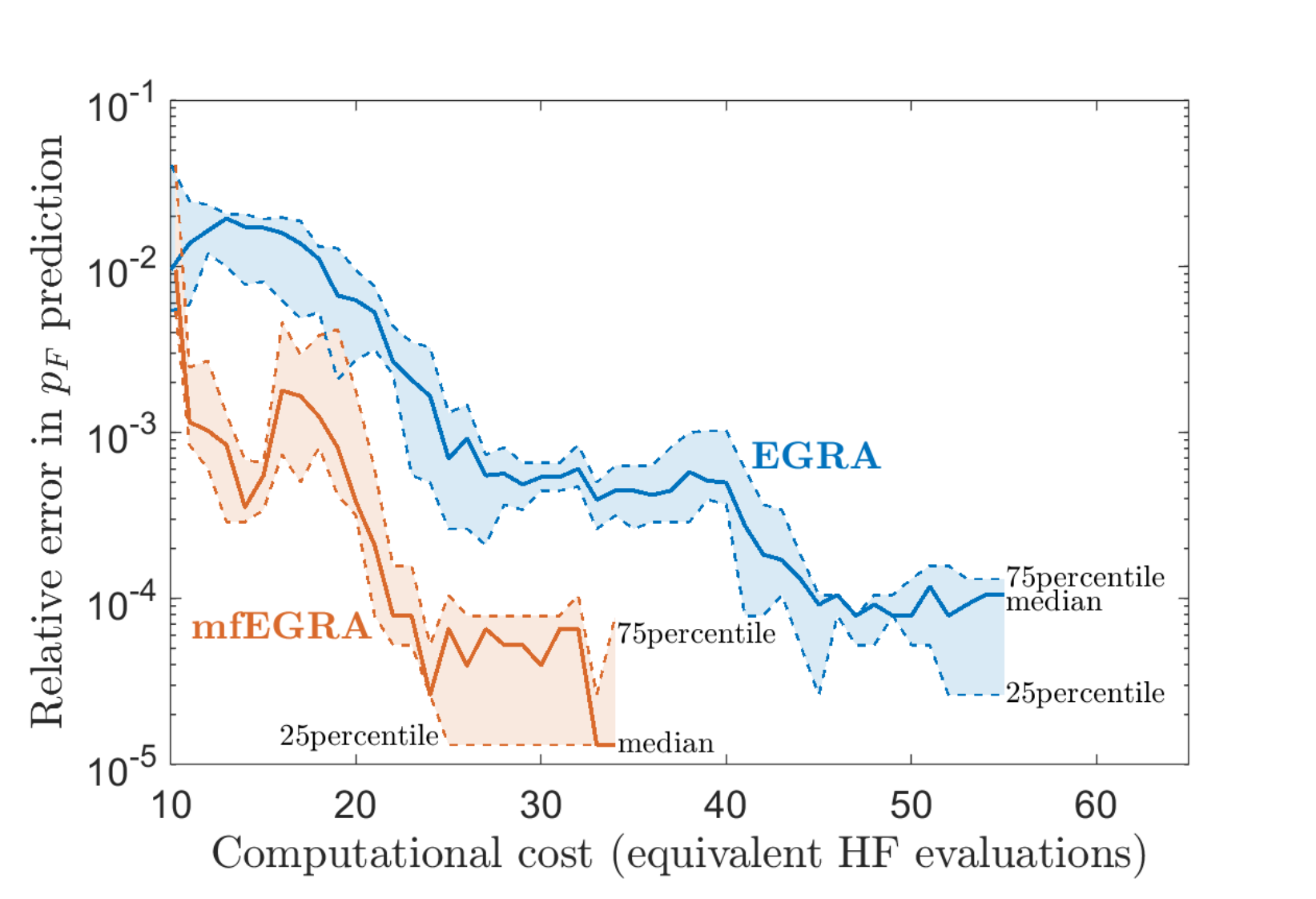}
	\caption{Comparing relative error in the estimate of probability of failure (shown in log-scale) using mfEGRA and single-fidelity EGRA for the three-dimensional acoustic horn application with 10 different initial DOEs.}
	\label{fig:Achorn_pf}
\end{figure}

\begin{figure}[!hbt]
	\centering
	\subfigure[]{\includegraphics[width=0.5\textwidth, trim = {0cm 0cm 1cm 0.5cm},clip,page=2]{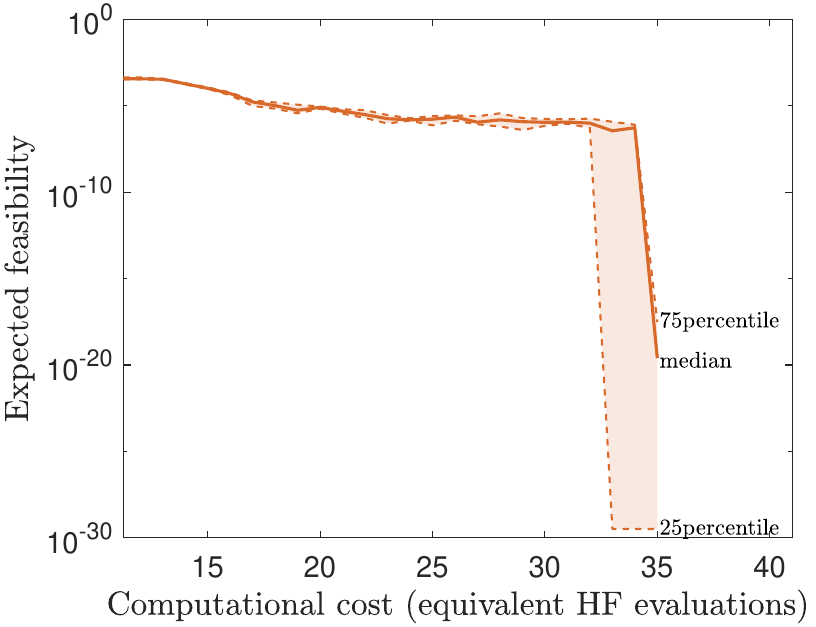}}\hfill
	\subfigure[]{\includegraphics[width=0.5\textwidth, trim = {0cm 0cm 1cm 0.5cm},clip,page=3]{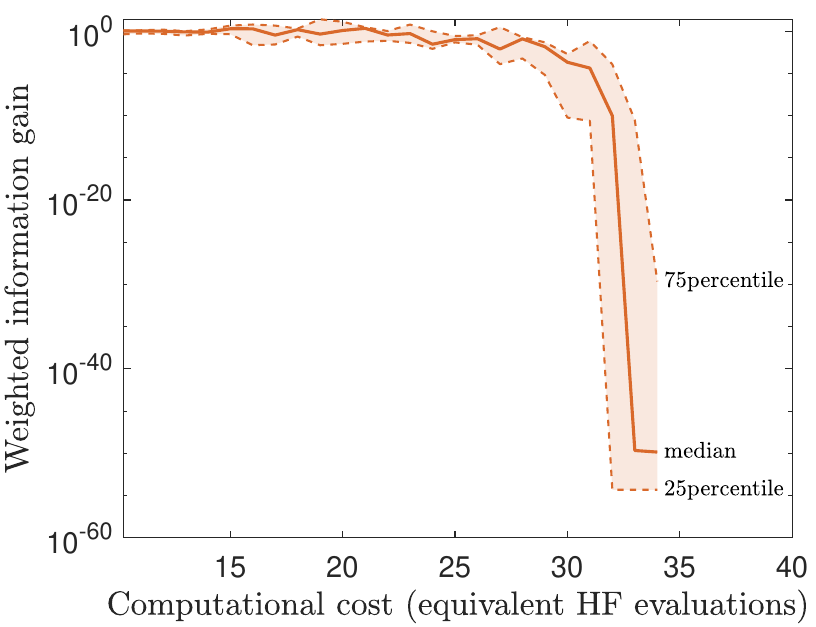}}
	\caption{Evolution of adaptive sampling criteria (a) expected feasibility function, and (b) weighted information gain for the three-dimensional acoustic horn application with 10 different initial DOEs.}
	\label{fig:AchornEF_IG}
\end{figure}

The reference probability of failure is estimated to be $p_\tF = 0.3812$ using $10^5$ Monte Carlo samples of the high-fidelity model. We repeat the mfEGRA and the single-fidelity EGRA results using 10 different initial DOEs with 10 samples in each (generated using Latin hypercube sampling) to get the confidence bands on the results. The comparison of convergence of the relative error in the probability of failure is shown in Figure~\ref{fig:Achorn_pf} for mfEGRA and single-fidelity EGRA. In this case, mfEGRA needs 19 equivalent high-fidelity solves to reach a median relative error value of below $10^{-3}$ as compared to 25 required by single-fidelity EGRA leading to 24\% reduction in computational cost. The reduction in computational cost using mfEGRA is driven by the discrepancy between the models and the relative cost of evaluating the models. In the acoustic horn case, we see computational savings of 24\% as compared to around 46\% seen in the analytic test problem in Section~\ref{s:test}. This can be explained by the substantial difference in relative costs -- 40 times cheaper low-fidelity model for the acoustic horn problem as compared to two low-fidelity models that are 100-1000 times cheaper than the high-fidelity model for the analytic test problem. The evolution of the mfEGRA adaptive sampling criteria can be seen in Figure~\ref{fig:AchornEF_IG}.
%, which can be used to define a tuned stopping criterion for the acoustic horn problem as mentioned before.
%

\begin{figure}[!hbt]
	\centering
	\subfigure[]{\includegraphics[width=0.5\textwidth, trim = {0cm 0cm 1cm 0cm},clip,page=4]{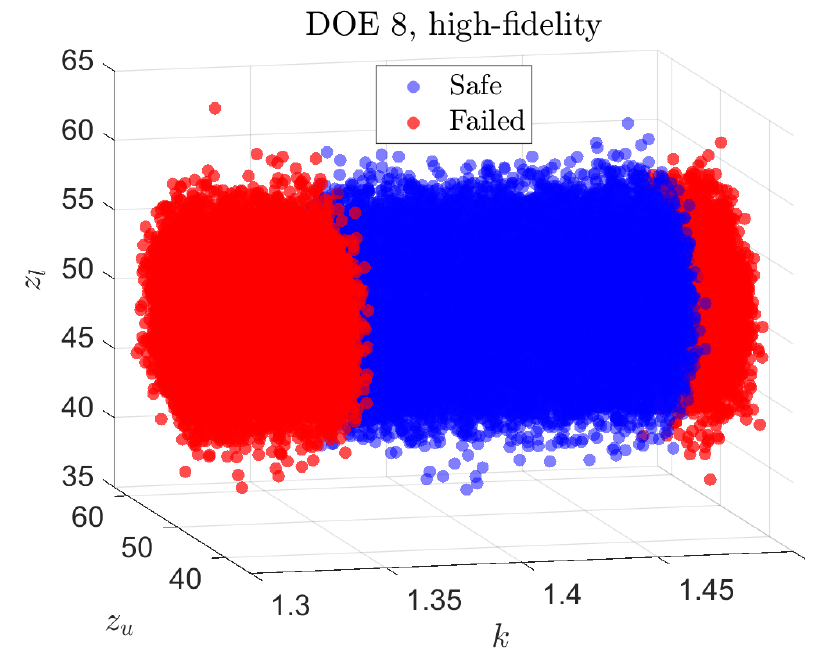}}\hfill
	\subfigure[]{\includegraphics[width=0.5\textwidth, trim = {0cm 0cm 1cm 0cm},clip,page=5]{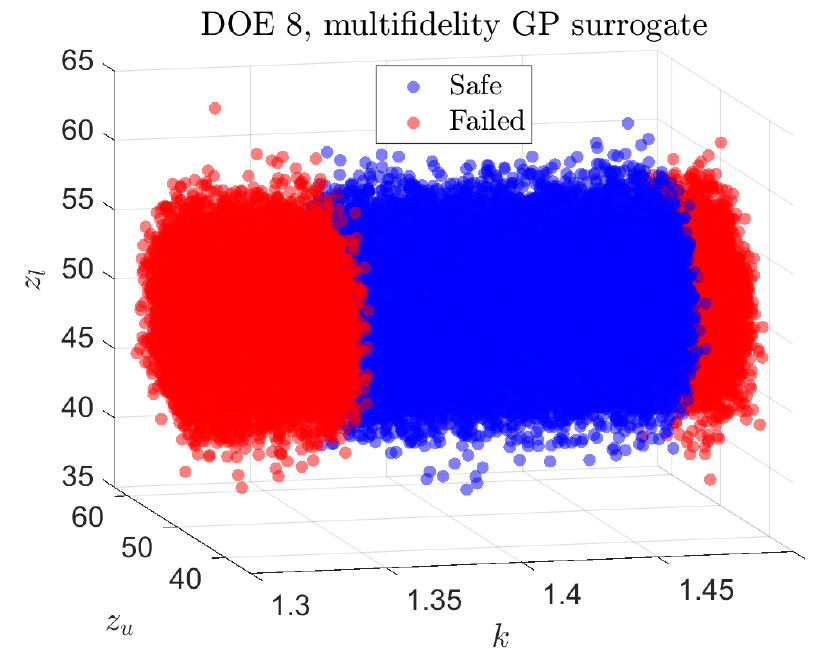}}
	\caption{Classification of Monte Carlo samples using (a) high-fidelity model, and (b) the final refined multifidelity GP surrogate for a particular initial DOE using mfEGRA for the three-dimensional acoustic horn problem.}
	\label{fig:AchornClassify}
\end{figure}

\begin{figure}[!hbt]
	\centering
	\includegraphics[width=\textwidth, trim = {1.5cm 0cm 1.5cm 0cm},clip,page=6]{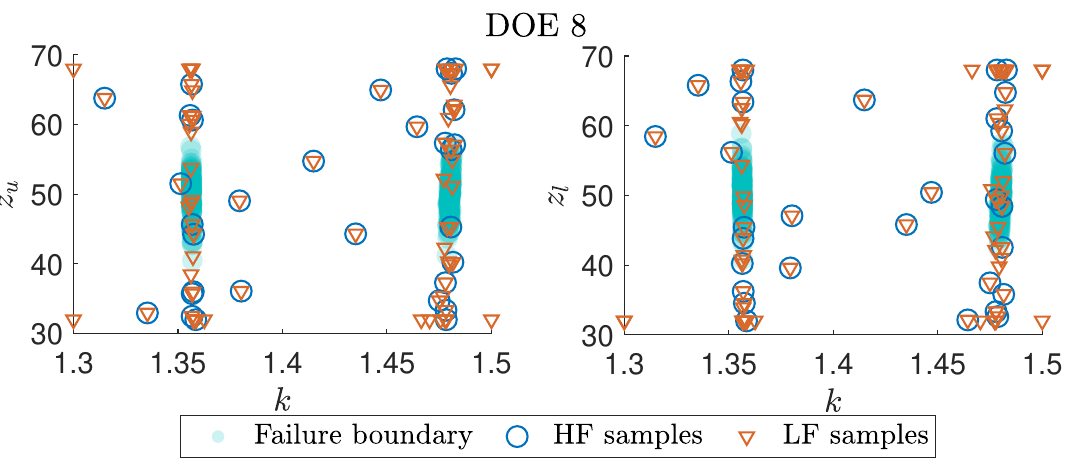}\hfill
	\caption{Location of samples from different fidelity models using mfEGRA for the three-dimensional acoustic horn problem for a particular initial DOE. The cloud of points are the high-fidelity Monte Carlo samples near the failure boundary.}
	\label{fig:AchornMFsamp}
\end{figure}

Figure~\ref{fig:AchornClassify} shows that classification of the Monte Carlo samples using the high-fidelity model and the adaptively refined surrogate model for a particular initial DOE lead to very similar results. It also shows that in the acoustic horn application there are two disjoint failure regions and the method is able to accurately capture both failure regions. The location of the samples from the different models when mfEGRA is used to refine the multifidelity GP surrogate for a particular initial DOE can be seen in Figure~\ref{fig:AchornMFsamp}. The figure shows that most of the high-fidelity samples are selected around the failure boundary. For this DOE, mfEGRA requires 31 evaluations of the high-fidelity model and 76 evaluations of the low-fidelity model to reach an EFF value below $10^{-10}$.

\begin{figure}[!hbt]
	\centering
	\includegraphics[width=12cm, trim = {0cm 0cm 1cm 1cm},clip,page=7]{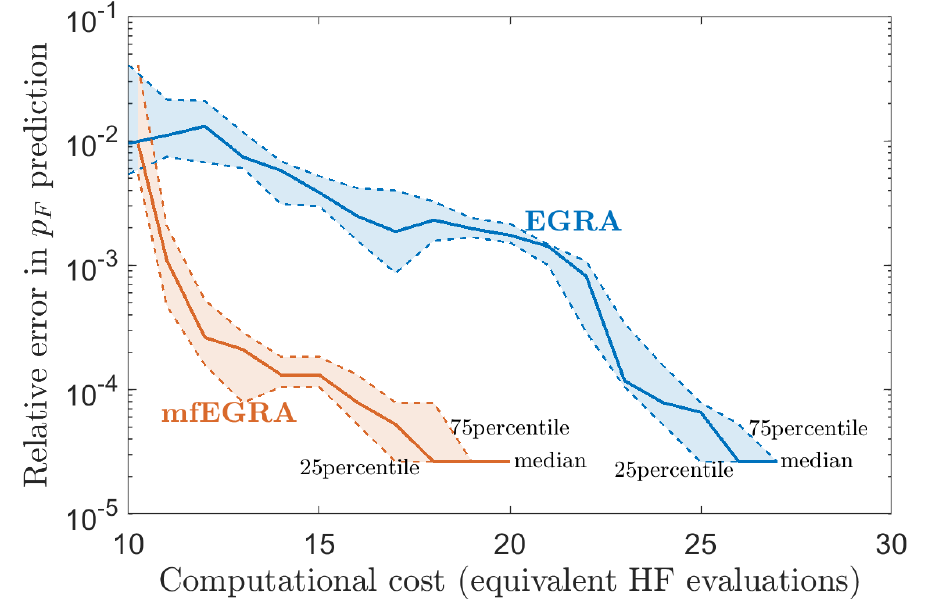}
	\caption{Comparing relative error in the estimate of probability of failure (shown in log-scale) using mfEGRA and single-fidelity EGRA by limiting the search space for adaptive sampling location to a set of Monte Carlo samples drawn from the given random variable distribution for the three-dimensional acoustic horn application with 10 different initial DOEs.}
	\label{fig:Achorn_pf_MC}
\end{figure}

Similar to the work in Refs.~\cite{echard2011ak,bect2012sequential}, EGRA and mfEGRA can also be implemented by limiting the search space for adaptive sampling location in Equation~\eqref{e:Loc} to the set of Monte Carlo samples (here, $10^5$) drawn from the given random variable distribution. The convergence of relative error in probability of failure estimate using this method improves for both mfEGRA and single-fidelity EGRA as can be seen in Figure~\ref{fig:Achorn_pf_MC}. In this case, mfEGRA requires 12 equivalent high-fidelity solves as compared to 22 high-fidelity solves required by single-fidelity EGRA to reach a median relative error below $10^{-3}$ leading to computational savings of around 45\%.

\subsubsection{Four-dimensional case}
For the four-dimensional acoustic horn problem, the inputs to the system are the three random variables used before along with a random variable $\xi$ defined by a truncated normal distribution representing manufacturing uncertainty as listed in Table~\ref{t:randvar2}. The parameters defining the geometry of the acoustic horn (see Figure~\ref{fig:AchornIllus}) are now given by $b_i+\xi, i=1,\dots,6$ to account for manufacturing uncertainty. In this case, we define failure of the system to be $s(\bz)>0.16$ to make the failure a rarer event. The limit state function is defined as $g(\bz) = s(\bz)-0.16$, which defines the failure boundary as $g(\bz)=0$. The reference probability of failure is estimated to be $p_\tF=7.2\times10^{-3}$ using $10^5$ Monte Carlo samples of the high-fidelity model. Note that the $p_\tF$ in the four-dimensional case is two orders of magnitude lower than the three-dimensional case. The complexity of the problem increases because of the higher dimensionality as well as the rarer event probability of failure to be estimated.
\begin{table}[!htb]
	\centering
	\caption{Random variables used in the four-dimensional acoustic horn problem.}
	\label{t:randvar2}
	\begin{tabular}{C{2cm}L{3cm}C{2cm}C{1cm}C{1cm}C{1cm}C{2cm}}
		\hline
		Random variable & Description & Distribution & Lower bound & Upper bound & Mean & Standard deviation \\ 
		\hline
		$k$ & wave number & Uniform & 1.3 & 1.5 & -- & --  \\
		$Z_u$ & upper horn wall impedance & Normal & -- & -- & 50 & 3 \\
		$Z_l$ & lower horn wall impedance & Normal & -- & -- & 50 & 3  \\
		$\xi$ & manufacturing uncertainty & Truncated Normal & -0.1 & 0.1 & 0 & 0.05  \\
		\hline
	\end{tabular}
\end{table}

\begin{figure}[!hbt]
	\centering
	\includegraphics[width=12cm, trim = {0cm 0cm 1cm 1cm},clip,page=1]{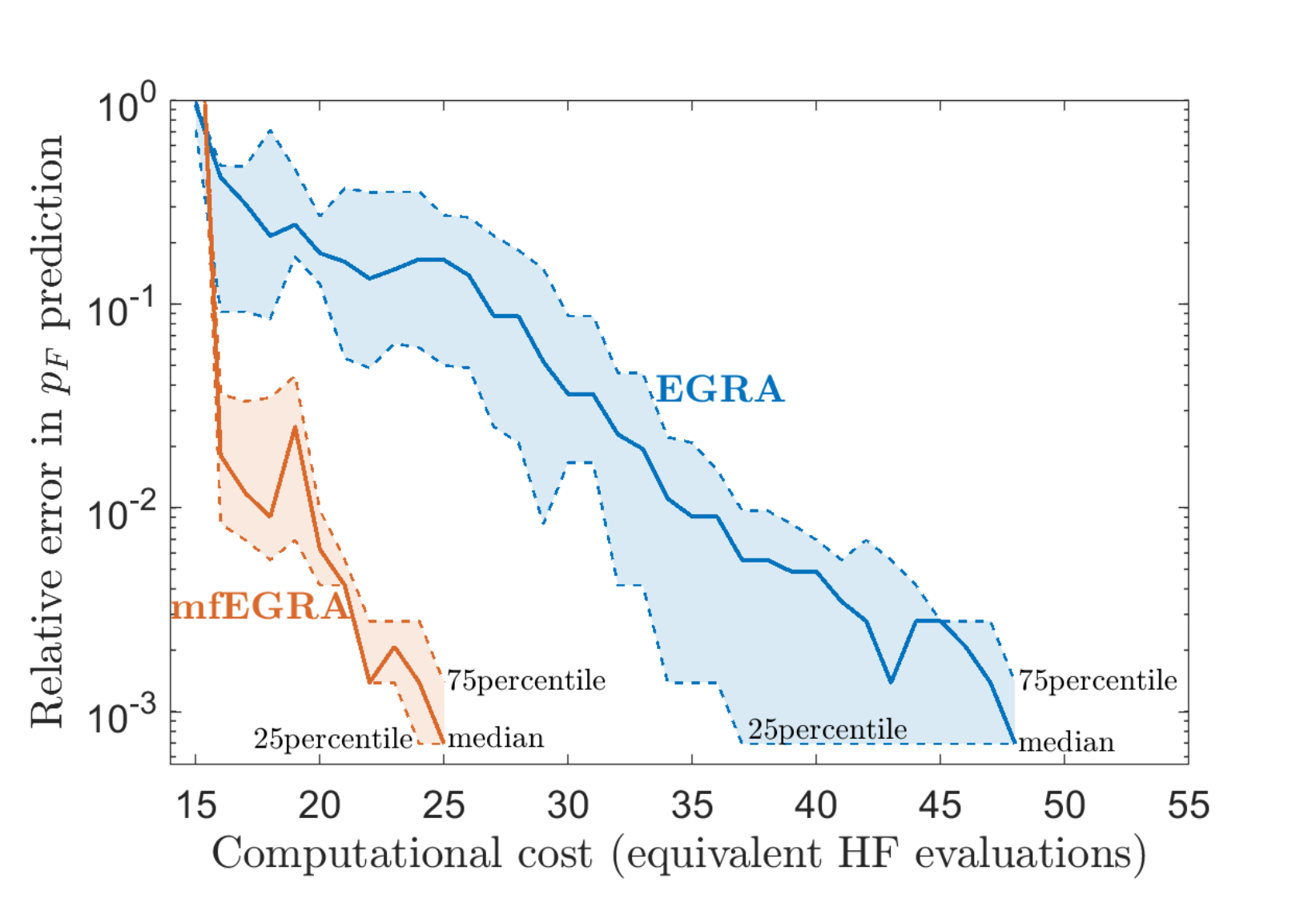}
	\caption{Comparing relative error in the estimate of probability of failure (shown in log-scale) using mfEGRA and single-fidelity EGRA by limiting the search space for adaptive sampling location to a set of Monte Carlo samples drawn from the given random variable distribution for the four-dimensional acoustic horn application with 10 different initial DOEs.}
	\label{fig:Achorn4D_pf_MC}
\end{figure}

In this case, we present the results for EGRA and mfEGRA implemented by limiting the search space to \textit{a priori} Monte Carlo samples (here, $10^5$) drawn from the given random variable distribution. Note that the lower probability of failure estimation required here necessitates the use of \textit{a priori} drawn Monte Carlo samples to efficiently achieve the required accuracy. The computational efficiency can be further improved by combining EGRA and mfEGRA with Monte Carlo variance reduction techniques, especially for problems with even lower probabilities of failure. We repeat the mfEGRA and the single-fidelity EGRA results using 10 different initial DOEs with 15 samples in each (generated using Latin hypercube sampling) to get the confidence bands on the results. The comparison of convergence of the relative error in the probability of failure is shown in Figure~\ref{fig:Achorn4D_pf_MC} for mfEGRA and single-fidelity EGRA. In this case, mfEGRA requires 25 equivalent high-fidelity solves as compared to 48 high-fidelity solves required by single-fidelity EGRA to reach a median relative error below $10^{-3}$ leading to computational savings of around 48\%.

%%%%%%%%%%%%%%%%%%%%%%%%%%%%%%%%%%%
\section{Concluding remarks}\label{s:5}
This paper introduces the mfEGRA (multifidelity EGRA) method that refines the surrogate to accurately locate the limit state function failure boundary (or any contour) while leveraging multiple information sources with different fidelities and costs. The method selects the next location based on the expected feasibility function and the next information source based on a weighted one-step lookahead information gain criterion to refine the multifidelity GP surrogate of the limit state function around the failure boundary.

We show through three numerical examples that mfEGRA efficiently combines information from different models to reduce computational cost. The mfEGRA method leads to computational savings of $\sim$46\% for a multimodal test problem and 24\% for a three-dimensional acoustic horn problem over the single-fidelity EGRA method when used for estimating the probability of failure. The mfEGRA method when implemented by restricting the search-space to \textit{a priori} drawn Monte Carlo samples showed even more computational efficiency with 45\% reduction in computational cost compared to single-fidelity method for the three-dimensional acoustic horn problem. We see that using \textit{a priori} drawn Monte Carlo samples improves the efficiency of both EGRA and mfEGRA, and the importance is further highlighted through the four-dimensional implementation of the acoustic horn problem, which requires estimating a rarer event probability of failure. For the four-dimensional acoustic horn problem, mfEGRA leads to computational savings of 48\% as compared to the single-fidelity method.
The driving factors for the reduction in computational cost for the method are the discrepancy between the high- and low-fidelity models, and the relative cost of the low-fidelity models compared to the high-fidelity model. These information are directly encoded in the mfEGRA adaptive sampling criterion helping it make the most efficient decision.

%%%%%%%%%%%%%%%%%%%%%%%%%%%%%%%%%%%%%%%%
\section*{Acknowledgements}
This work has been supported in part by the Air Force Office of Scientific Research (AFOSR) MURI on managing multiple information sources of multi-physics systems award numbers FA9550-15-1-0038 and FA9550-18-1-0023, the Air Force Center of Excellence on multi-fidelity modeling of rocket combustor dynamics award FA9550-17-1-0195, and the Department of Energy Office of Science AEOLUS MMICC award DE-SC0019303.

\bibliographystyle{aiaa}
\bibliography{MFEGRA_IG_bib}

\end{document}